\documentclass[10pt,twocolumn,letterpaper]{article}

\usepackage[pagenumbers]{cvpr} 

\usepackage[dvipsnames]{xcolor}

\newcommand{\parag}[1]{\smallskip\noindent\textbf{#1}\enspace}
\newcommand{\paragnoskip}[1]{\noindent\textbf{#1}\enspace}

\definecolor{C0}{HTML}{4C72B0}
\definecolor{C1}{HTML}{DD8452}
\definecolor{C2}{HTML}{55A868}
\definecolor{C3}{HTML}{C44E52}
\definecolor{C4}{HTML}{8172B3}
\definecolor{C5}{HTML}{937860}

\usepackage{multirow}
\newcommand{\method}{FAMix\xspace}
\usepackage{xcolor}
\usepackage{colortbl}
\usepackage{graphicx}
\usepackage{subcaption}

\usepackage{tikz}

\usepackage{stmaryrd}
\usetikzlibrary{arrows.meta}

\usepackage{amsmath}
\usepackage{amssymb}
\usepackage{pifont}
\newcommand{\xmark}{\ding{55}}
\usepackage{enumitem}
\usepackage{array}
\usepackage{makecell}
\usepackage[accsupp]{axessibility}
\usepackage[nice]{nicefrac}

\usepackage[most]{tcolorbox}
\usepackage{pgfplots}
\pgfplotsset{compat=newest}
\usepgfplotslibrary{groupplots}
\usepgfplotslibrary{dateplot}

\pgfplotsset{compat=1.11,
    /pgfplots/ybar legend/.style={
    /pgfplots/legend image code/.code={%
       \draw[##1,/tikz/.cd,yshift=-0.25em]
        (0cm,0cm) rectangle (3pt,0.8em);},
   },
}

\usepackage{tikz}

\usetikzlibrary{calc}
\usetikzlibrary{positioning}
\usetikzlibrary{angles,quotes}
\usetikzlibrary{backgrounds}
\usetikzlibrary{fit}
\usetikzlibrary{arrows}
\usetikzlibrary{arrows.meta}
\usetikzlibrary{shapes.symbols}
\usetikzlibrary{shadings}
\usetikzlibrary{shapes}
\usetikzlibrary{fadings}
\usetikzlibrary{bayesnet}
\usetikzlibrary{matrix}
\usetikzlibrary{plotmarks}
\usetikzlibrary{intersections}
\usetikzlibrary{pgfplots.fillbetween}
\pgfplotsset{compat=1.14}

\usepackage[linesnumbered,ruled,vlined]{algorithm2e}
\usepackage{algpseudocode}
\SetKwInput{KwInput}{Input}

\newcommand{\mb}[1]{\ensuremath{\mathbf{#1}}}
\newcommand{\mc}[1]{\ensuremath{\mathcal{#1}}}
\newcommand{\bs}[1]{\ensuremath{\boldsymbol{#1}}}

\newcommand{\src}[1]{{#1}_\text{s}}
\newcommand{\trg}[1]{{#1}_\text{t}}

\newcommand{\CLIPImOne}{\ensuremath{\texttt{CLIP-I1}}}
\newcommand{\CLIPImTwo}{\ensuremath{\texttt{CLIP-I2}}}
\newcommand{\CLIPText}{\ensuremath{\texttt{CLIP-T}}}

\definecolor{csroad}{RGB}{128, 64, 128}
\definecolor{csside}{RGB}{244, 35, 232}
\definecolor{csbuild}{RGB}{70, 70, 70}
\definecolor{cswall}{RGB}{102, 102, 156}
\definecolor{csfence}{RGB}{190, 153, 153}
\definecolor{cspole}{RGB}{153, 153, 153}
\definecolor{cslight}{RGB}{250, 170, 30}
\definecolor{cssign}{RGB}{220, 220, 0}
\definecolor{csveg}{RGB}{107, 142, 35}
\definecolor{csterrain}{RGB}{152, 251, 152}
\definecolor{cssky}{RGB}{70, 130, 180}
\definecolor{csperson}{RGB}{220, 20, 60}
\definecolor{csrider}{RGB}{255, 0, 0}
\definecolor{cscar}{RGB}{0, 0, 142}
\definecolor{cstruck}{RGB}{0, 0, 70}
\definecolor{csbus}{RGB}{0, 60, 100}
\definecolor{cstrain}{RGB}{0, 80, 100}
\definecolor{csmbike}{RGB}{0, 0, 230}
\definecolor{csbike}{RGB}{119, 11, 32}
\definecolor{csignore}{RGB}{0,0,0}

\definecolor{ours}{RGB}{220,220,220}

\definecolor{cvprblue}{rgb}{0.21,0.49,0.94}
\usepackage[pagebackref,breaklinks,colorlinks,citecolor=cvprblue]{hyperref}

\title{A Simple Recipe for Language-guided Domain Generalized Segmentation
    }

\author{Mohammad Fahes${^1}$ \quad Tuan-Hung Vu$^{1,2}$ \quad  Andrei Bursuc$^{1,2}$ \quad Patrick Pérez$^{3}$ \quad  Raoul de Charette$^{1}$
{}\\
$^1$ Inria \quad \quad \quad  $^2$ Valeo.ai \quad \quad \quad $^3$ Kyutai\\
}

\begin{document}
\twocolumn[{
\renewcommand\twocolumn[1][]{#1}
\maketitle
\thispagestyle{empty}
\begin{center}
\large
  \vspace{-1.5em}
  \centerline{
    \url{https://astra-vision.github.io/FAMix}
    }
    \vspace{-0.1cm}
 \end{center}
}]
\begin{abstract}
Generalization to new domains not seen during training is one of the long-standing challenges in deploying neural networks in real-world applications.
Existing generalization techniques either necessitate external images for augmentation, and/or aim at learning invariant representations by imposing various alignment constraints.
Large-scale pretraining has recently shown promising generalization capabilities, along with the potential of binding different modalities. 
For instance, the advent of vision-language models like CLIP has opened the doorway for vision models to exploit the textual modality.
In this paper, we introduce a simple framework for generalizing semantic segmentation networks by employing language as the source of randomization. 
Our recipe comprises three key ingredients: (i) the preservation of the intrinsic CLIP robustness through minimal fine-tuning, (ii) language-driven local style augmentation, and (iii) randomization by locally mixing the source and augmented styles during training.
Extensive experiments report state-of-the-art results on various generalization benchmarks. Code is accessible at \url{https://github.com/astra-vision/FAMix}.
\end{abstract}    
\section{Introduction}
\label{sec:intro}

A prominent challenge associated with deep neural networks is their constrained capacity to generalize when confronted with shifts in data distribution. This limitation is rooted in the assumption of data being independent and identically distributed, a presumption that frequently proves unrealistic in real-world scenarios. For instance, in safety-critical applications like autonomous driving, it is imperative for a segmentation model to exhibit resilient generalization capabilities when dealing with alterations in lighting, variations in weather conditions, and shifts in geographic location, among other considerations.

To address this challenge, domain adaptation~\cite{tzeng2017adversarial,long2018conditional,ganin2016domain,vu2019advent,hoffman2018cycada,li2019bidirectional} has emerged; its core principle revolves around aligning the distributions of both the source and target domains. However, DA hinges on having access to target data, which may not always be available. Even when accessible, this data might not encompass the full spectrum of distributions encountered in diverse real-world scenarios.
Domain generalization~\cite{zhou2022domain,zhou2021domain,wang2022generalizing,li2018domain,xu2021fourier,li2018deep} overcomes this limitation by enhancing the robustness of models to arbitrary and previously unseen domains.

The training of segmentation networks is often backed by large-scale pretraining as initialization for the feature representation. Until now, to the best of our knowledge, domain generalization for semantic segmentation (DGSS) networks~\cite{pan2018two,choi2021robustnet,peng2022semantic,zhao2022style,wu2022siamdoge,kim2022pin,yang2023generalized,huang2023style,lee2022wildnet,Kim_2023_ICCV} are pretrained with ImageNet~\cite{deng2009imagenet}. The underlying concept is to \emph{transfer} the representations from the upstream task of classification to the downstream task of segmentation.

Lately, contrastive language image pretraining~(CLIP)
\cite{radford2021learning,jia2021scaling,zhai2022lit,zhai2023sigmoid}
has demonstrated that transferable visual representations could be learned from the sole supervision of loose natural language descriptions at \emph{very} large scale. Subsequently, a plethora of applications have been proposed using CLIP~\cite{radford2021learning}, including zero-shot semantic segmentation~\cite{li2022languagedriven,zhou2022extract}, image editing~\cite{kwon2022clipstyler}, transfer learning~\cite{rao2022denseclip,fahes2023poda}, open-vocabulary object detection~\cite{gu2022openvocabulary}, few-shot learning~\cite{zhou2022learning,zhou2022conditional} etc. A recent line of research proposes fine-tuning techniques to preserve the robustness of CLIP under distribution shift~\cite{wortsman2022robust,kumar2022finetuning,goyal2023finetune,shu2023clipood}, but they are limited to classification.

In this paper, we aim at answering the following question: \emph{How to leverage CLIP pretraining for enhanced domain generalization for semantic segmentation?}  
The motivation for rethinking DGSS with CLIP is twofold. On one hand, distribution robustness is a notable characteristic of CLIP~\cite{fang2022data}. On the other hand, the language modality offers an 
extra source of information compared to unimodal pretrained models.

A direct comparison of training two segmentation models under identical conditions but with different pretraining, \ie ImageNet \vs CLIP, shows that CLIP pretraining does not yield promising results. Indeed, \cref{tab:imagenet_vs_CLIP} shows that fine-tuning CLIP-initialized network performs worse than its ImageNet counterpart on out-of-distribution (OOD) data.
This raises doubts about the suitability of CLIP pretraining for DGSS and indicates that it is more prone to \emph{overfitting} the source distribution at the expense of degrading its original distributional robustness properties. Note that both models converge and achieve similar results on in-domain data. More details are provided in~\cref{sec:clip_vs_imgnet}. 

\begin{table}[!t]
	\setlength{\tabcolsep}{0.01\linewidth}
	\centering
  \resizebox{1.\linewidth}{!}{
	\begin{tabular}{c|cccccccc|c}
		\toprule
		  Pretraining & \texttt{C} & \texttt{B} & \texttt{M} & \texttt{S} & \texttt{AN} & \texttt{AS} & \texttt{AR} & \texttt{AF} & Mean \\
		\midrule
        ImageNet & \textbf{29.04} & \textbf{32.17} & \textbf{34.26} & \textbf{29.87} & \textbf{4.36} & \textbf{22.38} & \textbf{28.34} & \textbf{26.76} & \textbf{25.90}\\
		CLIP & 16.81 & 16.31 & 17.80 & 27.10 & 2.95 & 8.58 & 14.35 & 13.61 & 14.69\\
		\bottomrule
	\end{tabular}
  }
        \vspace{-0.2cm}
        \smallskip\caption{
	\textbf{Comparison of ImageNet and CLIP pretraining for out-of-distribution semantic segmentation.} The network is DeepLabv3+ with ResNet-50 as backbone. The models are trained on GTAV and the performance (mIoU \%) is reported on Cityscapes (\texttt{C}), BDD-100K (\texttt{B}), Mapillary (\texttt{M}), SYNTHIA (\texttt{S}), and ACDC Night (\texttt{AN}), Snow (\texttt{AS}), Rain (\texttt{AR}) and Fog (\texttt{AF}).
	}
        \vspace{-0.3cm}
\label{tab:imagenet_vs_CLIP}
\end{table}

This paper shows that we can prevent such behavior with a simple recipe involving minimal fine-tuning, language-driven style augmentation, and mixing. 
Our approach is coined \method,
for \emph{Freeze, Augment and Mix}.

It was recently argued that fine-tuning might distort the pretrained representations and negatively affect OOD generalization~\cite{kumar2022finetuning}. To maintain the integrity of the representation, one extreme approach is to entirely freeze the backbone. However, this can undermine representation adaptability and lead to subpar OOD generalization. 
As a middle-ground strategy balancing adaptation and feature preservation, we suggest minimal fine-tuning of the backbone, where a substantial portion remains frozen, and only the final layers undergo fine-tuning. 

\begin{figure}[!t]
	\newcommand{\rowqual}[1]{
		\includegraphics[width=1.\linewidth]{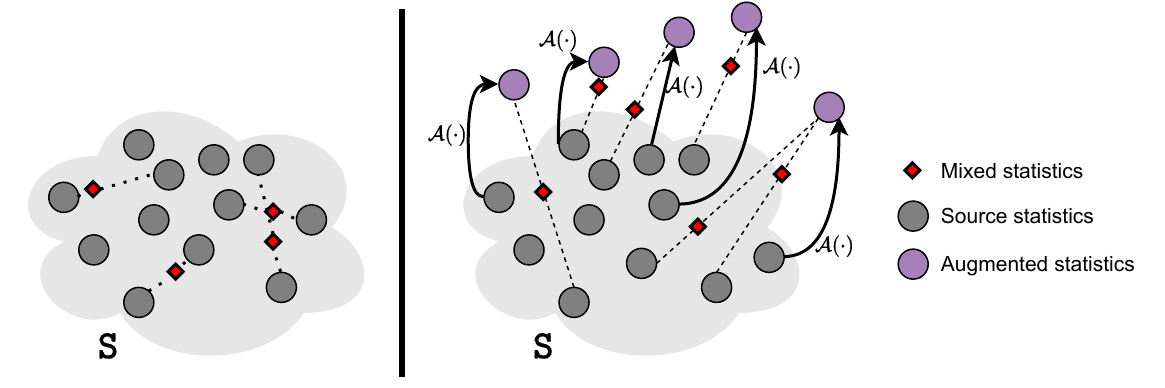}
	}
	\scriptsize	
	\setlength{\tabcolsep}{0.002\linewidth}
	\centering
		\begin{tabular}{c}
			\rowqual{}
		\end{tabular}
	\smallskip\caption{\textbf{Mixing strategies.} (\textit{Left}) MixStyle~\cite{zhou2021domain} consists of a linear mixing between the feature statistics of the source domain(s) $\textbf{S}$ samples. (\textit{Right}) We apply an augmentation $\mathcal{A(.)}$ on the source domain statistics, then perform linear mixing between original and augmented statistics. Intuitively, this enlarges the support of the training distribution by leveraging statistics beyond the source domain(s), as well as discovering intermediate domains. $\mathcal{A(.)}$ could be a language-driven or Gaussian noise augmentation, and we show that the former leads to better generalization results.}
\label{fig:aug_mix}
\end{figure}

For generalization, we show that rethinking MixStyle~\cite{zhou2021domain} leads to significant performance gains. 
As illustrated in~\cref{fig:aug_mix}, we mix the statistics of the original source features with augmented statistics mined using language. This helps explore styles beyond the source distribution at training time without 
using additional image.

We summarize our contributions as follows:
\begin{itemize}
    \item We propose a simple framework for DGSS based on minimal fine-tuning of the backbone and language-driven style augmentation. To the best of our knowledge, we are 
    the first to study DGSS with CLIP pretraining.
    \item We propose language-driven class-wise local style augmentation. We mine class-specific local statistics using prompts that express random styles and names of patch-wise dominant classes. During training, randomization is performed through patch-wise style mixing of the source and mined styles. 
    \item We conduct careful ablations to show the effectiveness of \method. Our framework outperforms state-of-the-art approaches in single and multi-source DGSS settings. 
\end{itemize}

\begin{figure*}[!t]
		\centering
		\includegraphics[width=0.95\linewidth]{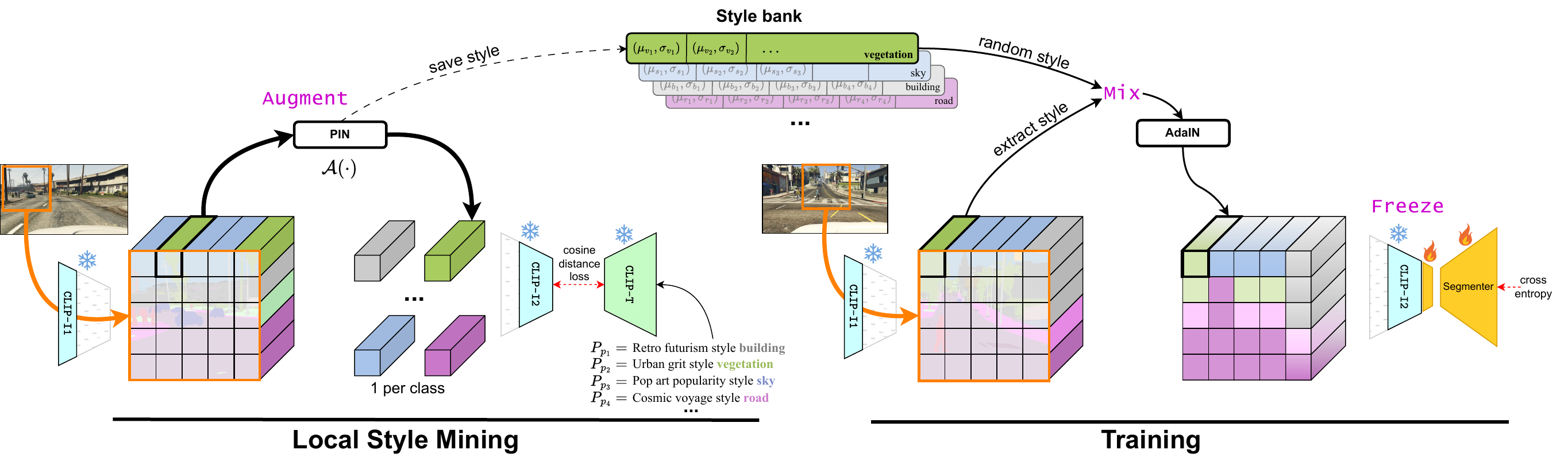}
	\smallskip\caption{\textbf{Overall process of \method.} \method consists of two steps. (\textit{Left}) \textit{Local style mining} consists of dividing the low-level feature activations into patches, which are used for style mining using Prompt-driven Instance Normalization (PIN)~\cite{fahes2023poda}. Specifically, for each patch, the dominant class is queried from the ground truth, and the mined style is added to corresponding class-specific style bank. (\textit{Right}) \textit{Training} the segmentation network is performed with minimal fine-tuning of the backbone. At each iteration, the low-level feature activations are viewed as grids of patches. For each patch, 
 the dominant class is queried using the ground truth, then a style is sampled from the corresponding style bank. Style randomization is performed by normalizing each patch in the grid by its statistics, and transferring the new style which is a mixing between the original style and the sampled one. The network is trained using only a cross-entropy loss.}
 \vspace{-0.2cm}
	\label{fig:method}
\end{figure*}

\section{Related works}
\label{sec:related}

\textbf{Domain generalization (DG).} The goal of DG is to train, from a single or multiple source domains, models that perform well under arbitrary domain shifts. The DG literature spans a broad range of approaches, including adversarial learning~\cite{li2018deep,zhao2020domain}, meta-learning~\cite{balaji2018metareg,qiao2020learning}, data augmentation~\cite{zhou2021domain,zhou2020deep,zhou2020learning} and domain-invariant representation learning~\cite{arjovsky2019invariant,ahuja2021invariance,krueger2021out,choi2021robustnet}. We refer the reader to~\cite{zhou2022domain,wang2022generalizing} for comprehensive surveys on DG.

\parag{Domain generalization with CLIP.} CLIP~\cite{radford2021learning} exhibits a remarkable distributional robustness~\cite{fang2022data}. Nevertheless, fine-tuning comes at the expense of sacrificing generalization. 
Kumar \textit{et al.}~\cite{kumar2022finetuning} observe that full fine-tuning can distort the pretrained representation, and propose a two-stage strategy, consisting of training a linear probe with a frozen feature extractor, then fine-tuning both. Wortsman \textit{et al.}~\cite{wortsman2022robust} propose ensembling the weights of zero-shot and fine-tuned models. Goyal \textit{et al.}~\cite{goyal2023finetune} show that preserving the pretraining paradigm (\ie contrastive learning) during the adaptation to the downstream task improves both in-domain (ID) and OOD performance without multi-step fine-tuning or weight ensembling. CLIPood~\cite{shu2023clipood} introduces margin metric softmax training objective and Beta moving average for optimization to handle both open-class and open-domain at test time.
On the other hand, distributional robustness  
could be improved by training a small amount of parameters on top of a frozen CLIP backbone in a teacher-student manner~\cite{jain2023efficiently, laroudie2023improving}. Other works show that specialized prompt ensembling and/or image ensembling strategies~\cite{allingham2023simple,ge2023improving} coupled with label augmentation using the WordNet hierarchy improve robustness in classification.

\parag{Domain Generalized Semantic Segmentation.}~DGSS methods could be categorized into three main groups: normalization methods, domain randomization (DR) and invariant representation learning. Normalization methods aim at removing style contribution from the representation. For instance, IBN-Net~\cite{pan2018two} shows that Instance Normalization (IN) makes the representation invariant to variations in the scene appearance (e.g., change of colors, illumination, etc.), and that combining IN and batch normalization (BN) helps the synthetic-to-real generalization. SAN \& SAW~\cite{peng2022semantic} proposes semantic-aware feature normalization and whitening, while RobustNet~\cite{choi2021robustnet} proposes an instance selective whitening loss, where only feature covariances that are sensitive to photometric transformations are whitened. DR aims instead at diversifying the data during training. Some methods use additional data for DR. For example, WildNet~\cite{lee2022wildnet} uses ImageNet~\cite{deng2009imagenet} data for content and style extension learning, while TLDR~\cite{Kim_2023_ICCV} proposes learning texture from random style images. Other methods like SiamDoGe~\cite{wu2022siamdoge} perform DR solely by data augmentation, using a Siamese~\cite{chen2021exploring} structure. 
Finally in the invariant representation learning group, SPC-Net~\cite{huang2023style} 
builds a representation space based on style and semantic projection and clustering, and SHADE~\cite{zhao2022style}
regularizes the training with a style consistency loss and a retrospection consistency loss.
\section{Method}
\label{sec:method}

\method proposes 
an effective
recipe for DGSS through the blending of simple ingredients. It consists of two stages (see \cref{fig:method}): 
(i) Local style mining from language  
(\cref{sec:local_sm}); (ii) Training of a segmentation network with minimal fine-tuning and local style mixing~(\cref{sec:train_famix}). 
In \cref{fig:method} and in the following, 
\CLIPImOne{} denotes the \textit{stem layers} and \textit{Layer1} of CLIP image encoder, \CLIPImTwo{} the remaining layers excluding the attention pooling, and \CLIPText{} the text encoder.

We start with some preliminary background knowledge, introducing AdaIN and PIN which are essential to our work.

\subsection{Preliminaries}
\label{sec:preli}

\paragnoskip{Adaptive Instance Normalization (AdaIN).} For a feature map $\mb{f} \in \mathbb{R}^{h \times w \times c}$, AdaIN~\cite{huang2017arbitrary} shows that the channel-wise mean $\bs{\mu} \in \mathbb{R}^{c}$ and standard deviation $\bs{\sigma} \in \mathbb{R}^{c}$ capture information about the style of the input image, allowing style transfer between images. Hence, stylizing a source feature $\src{\mb{f}}$ with an arbitrary target style $(\mu(\trg{\mb{f}}), \sigma(\trg{\mb{f}}))$ reads:
\begin{equation}
	\small
	\texttt{AdaIN}(\src{\mb{f}}, \trg{\mb{f}}) = \sigma(\trg{\mb{f}}) \Big( \frac{\src{\mb{f}} - \mu(\src{\mb{f}})}{\sigma(\src{\mb{f}})} \Big) + \mu(\trg{\mb{f}}),
	\label{eqn:adain}
\end{equation}
with $\mu(\cdot)$ and $\sigma(\cdot)$ the mean and standard deviation of input feature; multiplications and additions being element-wise.

\parag{Prompt-driven Instance Normalization (PIN).} PIN was introduced for prompt-driven zero-shot domain adaptation in P{\O}DA~\cite{fahes2023poda}. It replaces the target style $(\mu(\trg{\mb{f}}), \sigma(\trg{\mb{f}}))$ in AdaIN~\eqref{eqn:adain} with two optimizable variables $(\bs{\mu}, \bs{\sigma})$ 
guided by a single prompt in natural language. The rationale is to leverage a frozen CLIP~\cite{radford2021learning} to mine visual styles from the prompt representation in the shared space. Given a prompt $P$ and a feature map $\src{\mb{f}}$, PIN reads as:
\begin{equation}
	\small
	\texttt{PIN}_{(P)}(\src{\mb{f}}) = \bs{\sigma} \Big( \frac{\src{\mb{f}} - \mu(\src{\mb{f}})}{\sigma(\src{\mb{f}})} \Big) + \bs{\mu},
	\label{eqn:pin}
\end{equation}
where $\bs{\mu}$ and $\bs{\sigma}$ are optimized using gradient descent, such that the cosine distance between the visual feature representation and the prompt representation is minimized.

Different from P{\O}DA which mines styles globally with a predetermined prompt describing the target domain, we make use of PIN to mine class-specific styles using local patches of the features, leveraging \textit{random style prompts}. Further, we show the effectiveness of incorporating the class name in the prompt for better style mining.

\subsection{Local Style Mining}
\label{sec:local_sm}

Our approach is to leverage PIN to mine class-specific style banks that are used for feature augmentation when training \method.
Given a set of cropped images $\src{\mc{I}}$, we encode them using 
\CLIPImOne{} to get a set of low-level features $\src{\mc{F}}$. Each batch $b$ of features $\src{\mb{f}} \in \src{\mc{F}}$ is cropped into $m$ patches, resulting in $b\times m$ patches $\mb{f}_p$, and associated ground-truth annotation $\mb{y}_p$, of size $\nicefrac{h}{\sqrt{m}} \times \nicefrac{w}{\sqrt{m}} \times c$.  

We aim at populating $K$ style banks, $K$ being the total number of classes. For a feature patch $\mb{f}_p$, we compute the dominant class from the corresponding label patch $\mb{y}_p$, and get its name $t_p$ from the predefined classes in the training dataset. Given a set of prompts 
 describing random styles $\mc{R}$, the target prompt $P_p$ is formed by concatenating a randomly sampled style prompt \textcolor{red}{$r$} from $\mc{R}$ and \textcolor{blue}{$t_p$} (\eg, \texttt{\textcolor{red}{retro\! futurism\! style\!} \textcolor{blue}{building}}). We show in the experiments (\cref{sec:ablate}) that our method is not very sensitive to the prompt design, yet our prompt construction works 
best.

The idea is to mine \textit{proxy} domains and explore intermediate ones in a class-aware manner (as detailed in~\cref{sec:train_famix}), which makes our work fundamentally different from~\cite{fahes2023poda}, that steers features towards a particular target style and corresponding domain, and better suited to generalization.

To handle the class imbalance problem, we simply select one feature patch $\mb{f}_p$ per class among the total $b \times m$ patches, as shown in \cref{fig:method}. Consequently, we apply PIN~\eqref{eqn:pin} to optimize the local styles to match the representations of their corresponding prompts, and use the mined styles to populate the corresponding style banks. The complete procedure is 
outlined in~\cref{algo:local_mining}.

The resulting style banks \{$\mc{T}^{(1)}, \cdots, \mc{T}^{(K)}$\} are used for \textit{domain randomization} during 
training.

\begin{algorithm}[h!]
	\small
	\SetAlgoLined
	\SetKwFunction{DC}{get-dominant-class}
	\SetKwFunction{Name}{get-name}
	\SetKwFunction{Mean}{mean}
	\SetKwFunction{Std}{std}
	\SetKwFunction{Crop}{crop-patch}
	\SetKwFunction{PIN}{PIN}
	\SetKwFunction{Concat}{concat}
	\SetKwFunction{Sample}{sample}
	\SetKwInOut{Input}{Input}
	\SetKwInOut{Output}{Output} 
	\SetKwInOut{Parameter}{Param}
	
	\Input{Set $\src{\mc{F}}$ of source features batches. \\
		Label set $\src{\mc{Y}}$ in $\src{\mc{D}}$. \\
		Set of random prompts $\mc{R}$ and class names $\mc{C}$.
	}
	\Parameter{Number of patches $m$. \\
		Number of classes $K$. \\
}
\Output{$K$ sets \{$\mc{T}^{(1)}, \cdots, \mc{T}^{(K)}$\} of class-wise augmented statistics.}

$\{\mc{T}^{(1)}, \cdots, \mc{T}^{(K)}\} \gets \emptyset $ \\

\ForEach{$(\src{\mb{f}} \in \src{\mc{F}}$, $\src{\mb{y}}\in \src{\mc{Y}})$}{%
		$\{\mb{y}_p\} \gets \Crop(\src{\mb{y}},m)$ \\
		$\{c_p\}, \{P_p\},\{f_{p}\}  \gets \emptyset $ \\
		
		\ForEach{$\mb{y}_p$ $\in$ \{$\mb{y}_p$\}}{
			$c_p \gets \DC(\mb{y}_p)$ \\
			\If{$c_p$ not in $\{c_p\}$}
			{
				$\{c_p\} \gets c_p$ \\
				\mbox{$\{P_p\} \gets \Concat(\Sample(\mc{R}),\Name(c_p))$} \\
				$\{f_{p}\} \gets f_p $}    
		}
		$\bs{\mu}^{(c_p)}, \bs{\sigma}^{(c_p)}, \mb{f}_p' \gets \PIN_{(P_p)}(\mb{f}_p)$ \\
		$\mc{T}^{(c_p)} \gets \mc{T}^{(c_p)} \cup \{(\bs{\mu}^{(c_p)}, \bs{\sigma}^{(c_p)})\}$
	}
	
	\caption{%
		Local Style Mining.}
	\label{algo:local_mining}
\end{algorithm}

\begin{algorithm}[t!]
  \small
  \SetAlgoLined
  \SetKwInOut{Input}{Input}
  \SetKwInOut{Parameter}{Param}
  \SetKwFunction{DC}{get-dominant-class}
  \SetKwFunction{Sample}{sample}
  \SetKwFunction{adain}{AdaIN}
  \SetKwFunction{beta}{Beta}
  \SetKwFunction{loss}{Loss}
  \SetKwFunction{ce}{cross-entropy}

  \Input{Set $\src{\mc{F}}$ of source features batches. \\
		Label set $\src{\mc{Y}}$ in $\src{\mc{D}}$. \\
        $K$ sets \{$\mc{T}^{(1)}, \cdots, \mc{T}^{(K)}$\} of class-wise \\ augmented statistics.
	}
   \Parameter{Number of patches $m$.}
   \ForEach{($\src{\mb{f}} \in \src{\mc{F}}$, $\src{\mb{y}}\in \src{\mc{Y}}$)}{%
   $\alpha \sim \beta(0.1,0.1)$ \\
    \For{$(i,j) \in [1,\sqrt{m}] \times [1,\sqrt{m}]$} {
    $c_p^{(ij)} \gets \DC(\src{\mb{y}}^{(ij)})$ \\
    $\bs{\mu}^{(ij)}, \bs{\sigma}^{(ij)} \gets \Sample (\mc{T}^{(c_p^{(ij)})})$ \\
    
    $\mu_{\textit{mix}} \gets (1-\alpha). \mu(\src{\mb{f}}^{(ij)})  + \alpha. \bs{\mu}^{(ij)} $ \\
    $\sigma_{\textit{mix}} \gets (1-\alpha). \sigma(\src{\mb{f}}^{(ij)})  + \alpha. \bs{\sigma}^{(ij)} $ \\
    $\src{\mb{f}}^{(ij)} \gets \adain(\src{\mb{f}}^{(ij)},\mu_{\textit{mix}}, \sigma_{\textit{mix}}) $ \\
    }
    $\src{\mb{\tilde{y}}} \gets \CLIPImTwo{}(\src{\mb{f}})$ \\
    $\loss = \ce(\src{\mb{\tilde{y}}}, \src{\mb{y}})$ \\
  }
	\caption{
 Training \method.}
\label{algo:training_FAMIX}
\end{algorithm}

\begin{table}[ht!]
    \small
	\setlength{\tabcolsep}{0.008\linewidth}
	\centering
	\resizebox{1.\linewidth}{!}{%
	\begin{tabular}{ll|cccccccc|c}
		\toprule
		  Method & arch. & \texttt{C} & \texttt{B} & \texttt{M} & \texttt{S} & \texttt{AN} &  \texttt{AS} &  \texttt{AR} & \texttt{AF} & Mean \\
		\midrule
		  RobustNet~\cite{choi2021robustnet} & \multirow{12}{*}{RN50} & 36.58 & 35.20 & 40.33 & 28.30 & 6.32 & 29.97 & 33.02 & 32.56 & 30.29\\
        SAN \& SAW~\cite{peng2022semantic} & & 39.75 & 37.34 & 41.86 & 30.79 & - & - & - & -& -\\
        Pin the memory~\cite{kim2022pin}  & &41.00	& 34.60 &	37.40 &	27.08	& 3.84 & 5.51 & 5.89 & 7.27 & 20.32\\ 
        SHADE~\cite{zhao2022style} & &44.65 &	39.28 &	43.34 & 28.41 & 8.18 & 30.38 & 35.44 & 36.87 &  33.32\\
        SiamDoGe~\cite{wu2022siamdoge} & & 42.96 & 	37.54 &	40.64	& 28.34& 	10.60& 	30.71&	35.84	& 36.45 & 32.89\\
        DPCL~\cite{yang2023generalized} & & 44.87&	40.21&	\underline{46.74}	&-	&-	&-	&-	&-&- \\ 
        SPC-Net~\cite{huang2023style} 
        & & 44.10&	40.46&	45.51&	-	&-	&-	&-	&-&-\\
        NP~\cite{fan2023towards} & & 40.62 & 35.56 & 38.92 & 27.65 & -  & - &-	&-&-\\ 
         WildNet*~\cite{lee2022wildnet}  & & 44.62 & 38.42 & 46.09 & \underline{31.34} & 8.27 & 30.29 & 36.32 & 35.39 & 33.84 \\
        TLDR*~\cite{Kim_2023_ICCV}  & & \underline{46.51} & \underline{42.58}	& 46.18	& 30.57 & \underline{13.13} & \underline{36.02} & \textbf{38.89} & \textbf{40.58} & \underline{36.81} \\ 

	   \rowcolor{ours} \method (ours) & & \textbf{48.15} & \textbf{45.61} & \textbf{52.11} & \textbf{34.23} & \textbf{14.96} & \textbf{37.09} & \underline{38.66} &  \underline{40.25} & \textbf{38.88}\\
  \midrule
        SAN \& SAW~\cite{peng2022semantic} & \multirow{5}{*}{RN101} & 45.33 & 41.18 & 40.77 & 31.84 & - & - & - & - & - \\ 
        SHADE$^\dagger$~\cite{zhao2022style} & & 46.66 & 43.66 & 45.50 & 31.58 & \underline{7.58} & \underline{32.48} & \underline{36.90} & \underline{36.69} & \underline{35.13} \\
        WildNet*~\cite{lee2022wildnet} &  & 45.79 & 41.73 & 47.08 & 32.51 & - & - & - & - & - \\
        TLDR*~\cite{Kim_2023_ICCV} &  & \underline{47.58} & \underline{44.88} & \underline{48.80} & \underline{33.14} &  - & - & - & - & - \\
		  \rowcolor{ours} \method (ours) & & \textbf{49.47} & \textbf{46.40} & \textbf{51.97} & \textbf{36.72} & \textbf{19.89} & \textbf{41.38} & \textbf{40.91} & \textbf{42.15} & \textbf{41.11}\\
    \bottomrule
	\end{tabular}%
	}
 \vspace{-0.1cm}
    \caption{
	\textbf{Single-source DGSS trained on GTAV.} Performance (mIoU \%) of \method compared to other DGSS methods trained on \texttt{G} and evaluated on \texttt{C}, \texttt{S}, \texttt{M}, \texttt{S}, \texttt{A} for ResNet-50 (`RN50') and ResNet-101 (`RN101') backbone architecture (`arch.'). * indicates the use of extra-data. $\dagger$ indicates the use of the full data for training. We emphasize \textbf{best} and \underline{second best} results.
	}
 \vspace{-0.3cm}
 \label{tab:compare_sota_gta5}
\end{table}

\subsection{Training \method}
\label{sec:train_famix}

\paragnoskip{Style randomization.} During training, randomly cropped images $\src{\mc{I}}$ are encoded into $\src{\mb{f}}$ using \CLIPImOne{}. Each batch of feature maps $\src{\mb{f}}$ is viewed as a grid of $m$ patches, without cropping them. For each patch $\src{\mb{f}}{\!}^{(ij)}$ within the grid, the dominant class $c_p{\!}^{(ij)}$ is queried using the corresponding ground truth patch $\src{\mb{y}}{\!}^{(ij)}$, and a style is randomly sampled from the corresponding mined bank $\mc{T}{}^{(c_p{\!}^{(ij)})}$. We then apply patch-wise
convex combination (\ie, style mixing) of the original style of the patch and the mined style. Specifically, for an arbitrary patch $\src{\mb{f}}{\!}^{(ij)}$, our local style mixing reads:
\begin{align}
    \mu_{\textit{mix}} \gets (1-\alpha) \mu(\src{\mb{f}}^{(ij)})  + \alpha \bs{\mu}^{(ij)} \label{eq:mu_mix}\\
    \sigma_{\textit{mix}} \gets (1-\alpha) \sigma(\src{\mb{f}}^{(ij)})  + \alpha \bs{\sigma}^{(ij)},
    \label{eq:sigma_mix}
\end{align}
with $(\bs{\mu}^{(ij)}, \bs{\sigma}^{(ij)}) \in \mc{T}^{(c_p^{(ij)})}$ and $\alpha \in [0,1]^c$. 

As shown in~\cref{fig:aug_mix}, our style mixing strategy differs from~\cite{zhou2021domain} which applies a linear interpolation between styles extracted from the images of a limited set of \textit{source} domain(s) assumed to be available for training. 
Here, we view the mined styles as variations of multiple \textit{proxy} target domains defined by the prompts. Training is conducted over all the paths in the feature space between the source and proxy domains without requiring any additional image during training other than the one from source.

Style transfer is applied through AdaIN~\eqref{eqn:adain}. Only the standard cross-entropy loss between the ground truth $\src{\mb{y}}$ and the prediction $\src{\mb{\tilde{y}}}$ is applied for training the network. \cref{algo:training_FAMIX} shows the training steps of \method.

\parag{Minimal fine-tuning.} During training, we fine-tune only the last few layers of the backbone.
Subsequently, we examine various alternatives and show that the minimal extent of fine-tuning is the crucial factor in witnessing the effectiveness of our local style mixing strategy. 

Previous works~\cite{zhou2021domain,fan2023towards,pan2018two} suggest that shallow feature statistics capture style information while deeper features encode semantic content. Consequently, some DGSS methods focus on learning style-agnostic representations~\cite{choi2021robustnet,pan2018two,peng2022semantic}, but this can compromise the 
expressiveness of the representation and suppress content information. In contrast, our intuition is to retain these identified traits by introducing variability to the shallow features through augmentation and mixing. Simultaneously, we guide the network to learn invariant high-level representations by training the final layers of the backbone with a label-preserving assumption, using a standard cross-entropy loss.
\section{Experiments}
\label{sec:exp}

\subsection{Experimental setup}
\label{sec:exp_setup}

\paragnoskip{Synthetic datasets.} GTAV~\cite{richter2016playing} and SYNTHIA~\cite{ros2016synthia} are used as synthetic datasets. GTAV consists of 24\,966 images split into 12\,403 images for training, 6\,382 for validation and 6\,181 for testing. SYNTHIA consists of 9\,400 images: 6\,580 for training and 2\,820 for validation. GTAV and SYNTHIA are denoted by \texttt{G} and \texttt{S}, respectively.

\parag{Real datasets.}
Cityscapes~\cite{cordts2016cityscapes}, BDD-100K~\cite{yu2020bdd100k}, and Mapillary~\cite{neuhold2017mapillary} contain 2\,975, 7\,000, and 18\,000 images for training and 500, 1\,000, and 2\,000 images for validation, respectively. ACDC~\cite{sakaridis2021acdc} is a dataset of driving scenes in adverse conditions: night, snow, rain and fog with respectively 106, 100, 100 and 100 images in the validation sets. \texttt{C}, \texttt{B}, and \texttt{M} denote Cityscapes, BDD-100K and Mapillary, respectively; \texttt{AN}, \texttt{AS}, \texttt{AR} and \texttt{AF} denote night, snow, rain and fog subsets of ACDC, respectively.

\parag{Implementation details.}
Following previous works~\cite{choi2021robustnet,peng2022semantic,kim2022pin,zhao2022style,wu2022siamdoge,yang2023generalized,huang2023style,lee2022wildnet,Kim_2023_ICCV}, we adopt DeepLabv3+~\cite{chen2018encoder} as segmentation model. \mbox{ResNet-50} and \mbox{ResNet-101}~\cite{he2016deep}, initialized with CLIP pretrained weights, are used in our experiments as backbones. Specifically, we remove the attention pooling layer and add a randomly initialized decoder head. The output stride is $16$. Single-source and multi-source models are trained respectively for $40K$ and $60K$ iterations with a batch size of $8$. The training images are cropped to $768\times768$. Stochastic Gradient Descent (SGD) with a momentum of $0.9$ and weight decay of $10^{-4}$ is used as optimizer. Polynomial decay with a power of $0.9$ is used, with an initial learning rate of $10^{-1}$ for the classifier and $10^{-2}$ for the backbone. We use color jittering and horizontal flip as data augmentation. Label smoothing regularization~\cite{szegedy2016rethinking} is adopted. For style mining, \textit{Layer1} features are divided into $9$ patches. Each patch is resized 
to $56\times56$,
corresponding to the dimensions of \textit{Layer1} features for an input image of size $224\times224$ (\ie the input dimension of CLIP). We use ImageNet templates\footnote{\url{https://github.com/openai/CLIP/}} for each prompt.

\parag{Evaluation metric.} We evaluate our models on the validation sets of the unseen target domains with mean Intersection over Union (mIoU\%) of the $19$ shared semantic classes. For each experiment, we report the average of \textit{three runs}. 

\subsection{Comparison with DGSS methods}
\label{sec:compare_sota}

\textbf{Single-source DGSS.} We compare \method with state-of-the-art DGSS methods under the single-source setting.

\smallskip\noindent\underline{Training on GTAV~(\texttt{G})} as source, \cref{tab:compare_sota_gta5} reports models trained with either \mbox{ResNet-50} or \mbox{ResNet-101} backbones. The unseen target datasets are \texttt{C}, \texttt{B}, \texttt{M}, \texttt{S}, and the four subsets of \texttt{A}. \cref{tab:compare_sota_gta5} shows that our method significantly outperforms all the baselines on all the datasets for both backbones. We note that WildNet~\cite{lee2022wildnet} and TLDR~\cite{Kim_2023_ICCV} use extra-data, while SHADE~\cite{zhao2022style} uses the full \texttt{G} dataset (24,966 images) for training with \mbox{ResNet-101}. Class-wise performances are reported in~\cref{sec:class_perf}.

\smallskip\noindent\underline{Training on Cityscapes~(\texttt{C})} as source, \cref{tab:cstrain} reports performance with \mbox{ResNet-50} backbone. The unseen target datasets are \texttt{B}, \texttt{M}, \texttt{G}, and \texttt{S}. The table shows that our method outperforms the baseline in average, and is competitive to SOTA on \texttt{G} and \texttt{M}.

\begin{table}[t!]
    \small
	\setlength{\tabcolsep}{0.025\linewidth}
	\centering
    \resizebox{0.98\linewidth}{!}{%
    \begin{tabular}{c|cccc|c}
        \toprule
        Method & \texttt{B} & \texttt{M} & \texttt{G} & \texttt{S} & Mean\\
        \midrule
        RobustNet~\cite{choi2021robustnet} & 50.73 & 58.64 & 45.00 & 26.20 & 45.14\\
        Pin\,the\,memory~\cite{kim2022pin} &  46.78 & 55.10 &  - & - & -\\
        SiamDoGe~\cite{wu2022siamdoge} & 51.53 & \textbf{59.00} & 45.08 & 26.67 & 45.57\\ 
        WildNet*~\cite{lee2022wildnet} & 50.94 & \underline{58.79} & \textbf{47.01} & \underline{27.95} & \underline{46.17}\\
        DPCL~\cite{yang2023generalized} & \underline{52.29} & - & \underline{46.00} & 26.60 & -\\
        \rowcolor{ours} FAMix (ours) & \textbf{54.07} & 58.72 & 45.12 & \textbf{32.67} & \textbf{47.65}\\
        \bottomrule
        \end{tabular}%
    }%
    \caption{\textbf{Single-source DGSS trained on Cityscapes.} Performance (mIoU \%) of \method compared to other DGSS methods trained on \texttt{C} and evaluated on \texttt{B}, \texttt{M}, \texttt{G} and \texttt{S} for \mbox{ResNet-50} backbone. * indicates the use of extra-data. We emphasize \textbf{best} and \underline{second best} results.}%
    \label{tab:cstrain}%
\end{table}%

\parag{Multi-source DGSS.} We also show the effectiveness of \method in the multi-source setting, training on \texttt{G}+\texttt{S} and evaluating on \texttt{C}, \texttt{B} and \texttt{M}. The results reported in~\cref{tab:compare_sota_multisource_rn50} for \mbox{ResNet-50} backbone outperform state-of-the-art.

\begin{table}[t!]
    \small
	\setlength{\tabcolsep}{0.04\linewidth}
	\centering
	\resizebox{0.98\linewidth}{!}{%
	\begin{tabular}{c|ccc|c}
		\toprule
		  Method & \texttt{C} & \texttt{B} & \texttt{M} & Mean \\
		\midrule
		  RobustNet~\cite{choi2021robustnet} &  37.69 & 34.09 & 38.49 & 36.76 \\
        Pin the memory~\cite{kim2022pin} &  44.51 &  38.07 & 42.70 & 41.76 \\
        SHADE~\cite{zhao2022style} & 47.43 & 40.30 & 47.60 & 45.11 \\ 
        SPC-Net~\cite{huang2023style} & 46.36 & \underline{43.18} & \underline{48.23} & 45.92\\
        TLDR*~\cite{Kim_2023_ICCV} & \underline{48.83} & 42.58 & 47.80 & \underline{46.40}\\
		  \rowcolor{ours} \method (ours) & \textbf{49.41} & \textbf{45.51} & \textbf{51.61} & \textbf{48.84}  \\
		\bottomrule
	\end{tabular}%
	}
    \caption{
	\textbf{Multi-source DGSS trained on GTAV + SYNTHIA.} Performance (mIoU \%) of \method compared to other DGSS methods trained on \texttt{G}+\texttt{S} and evaluated on \texttt{C}, \texttt{B}, \texttt{M} for \mbox{ResNet-50} backbone. * indicates the use of extra-data. We emphasize \textbf{best} and \underline{second best} results.
	}
\label{tab:compare_sota_multisource_rn50}
\end{table}

\begin{figure*}[h]
	\newcommand{\rowqual}[1]{ %
		\includegraphics[width=0.16\linewidth]{figures/qualitative/#1_image.png} &
        \includegraphics[width=0.16\linewidth]{figures/qualitative/#1_gt.png} & 
		\includegraphics[width=0.16\linewidth]{figures/qualitative/#1_pin_mem.png} &
	    \includegraphics[width=0.16\linewidth]{figures/qualitative/#1_shade.png}&
        \includegraphics[width=0.16\linewidth]{figures/qualitative/#1_wildnet.png}&
		\includegraphics[width=0.16\linewidth]{figures/qualitative/#1_pred.png}
	}
	\scriptsize	
	\setlength{\tabcolsep}{0.002\linewidth}
	\centering
        \tiny
	\begin{tabular}{ccccccc}
				& {\scriptsize Image} & {\scriptsize GT} &  {\scriptsize PIN the mem.~\cite{kim2022pin}} & {\scriptsize SHADE~\cite{zhao2022style}} & {\scriptsize WildNet~\cite{lee2022wildnet}} & {\scriptsize\method}\\
		    \multirow{1}{*}[3em]{\texttt{C}} & \rowqual{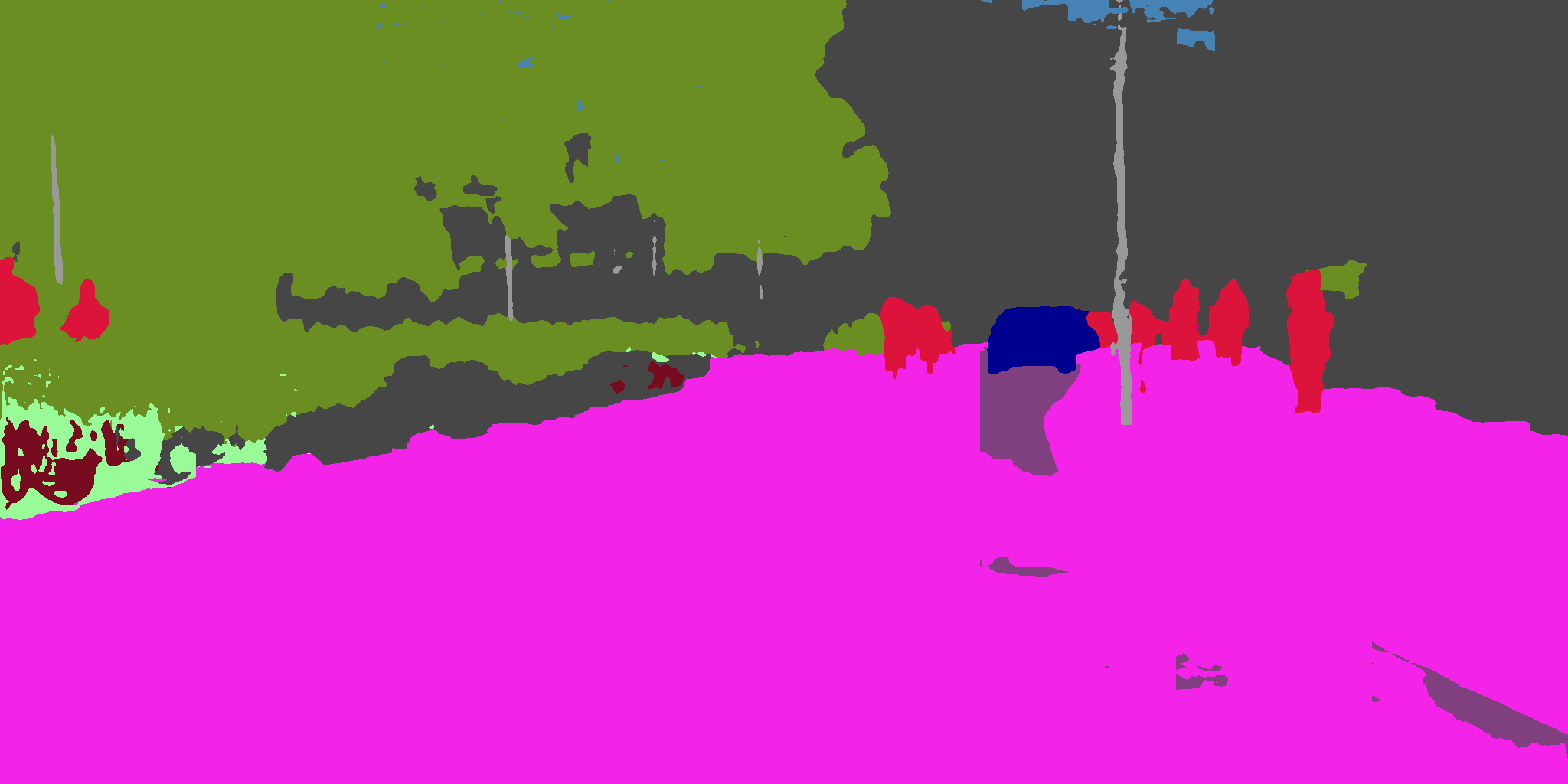}\\ 
		      \multirow{1}{*}[4em]{\texttt{B}} & \rowqual{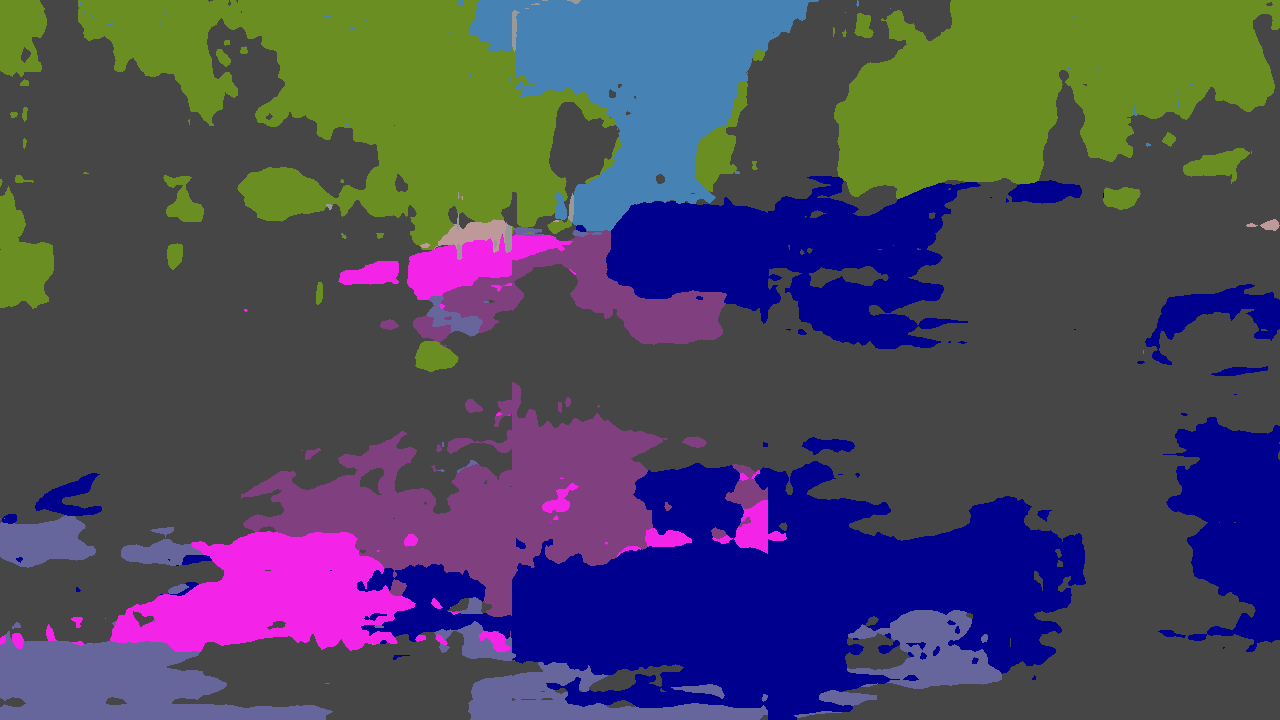}\\ 
            \multirow{1}{*}[4em]{\texttt{M}} & \includegraphics[width=0.16\linewidth,trim =0cm 6cm 0cm 15cm, clip]{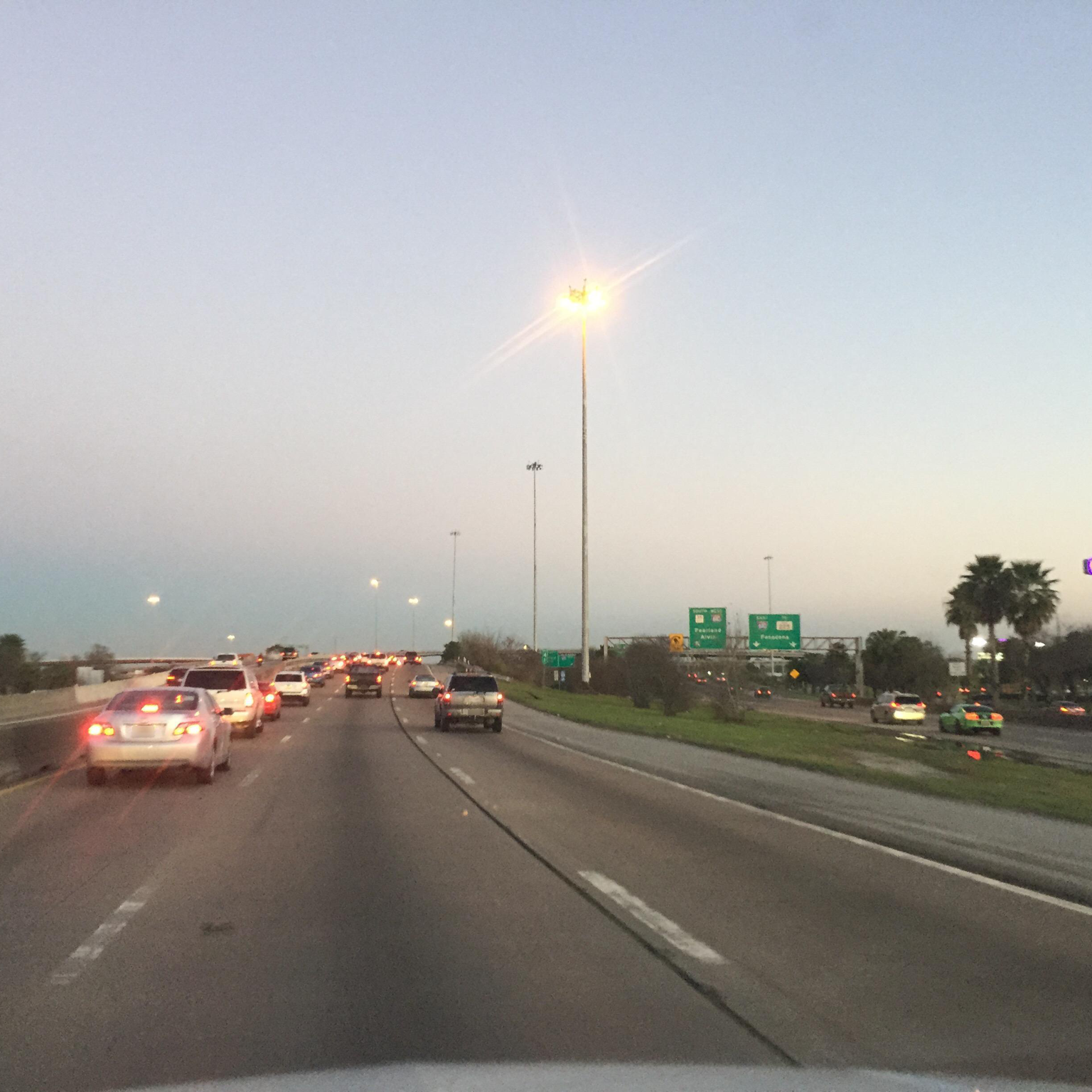} &
        \includegraphics[width=0.16\linewidth,trim=0cm 6cm 0cm 15cm, clip]{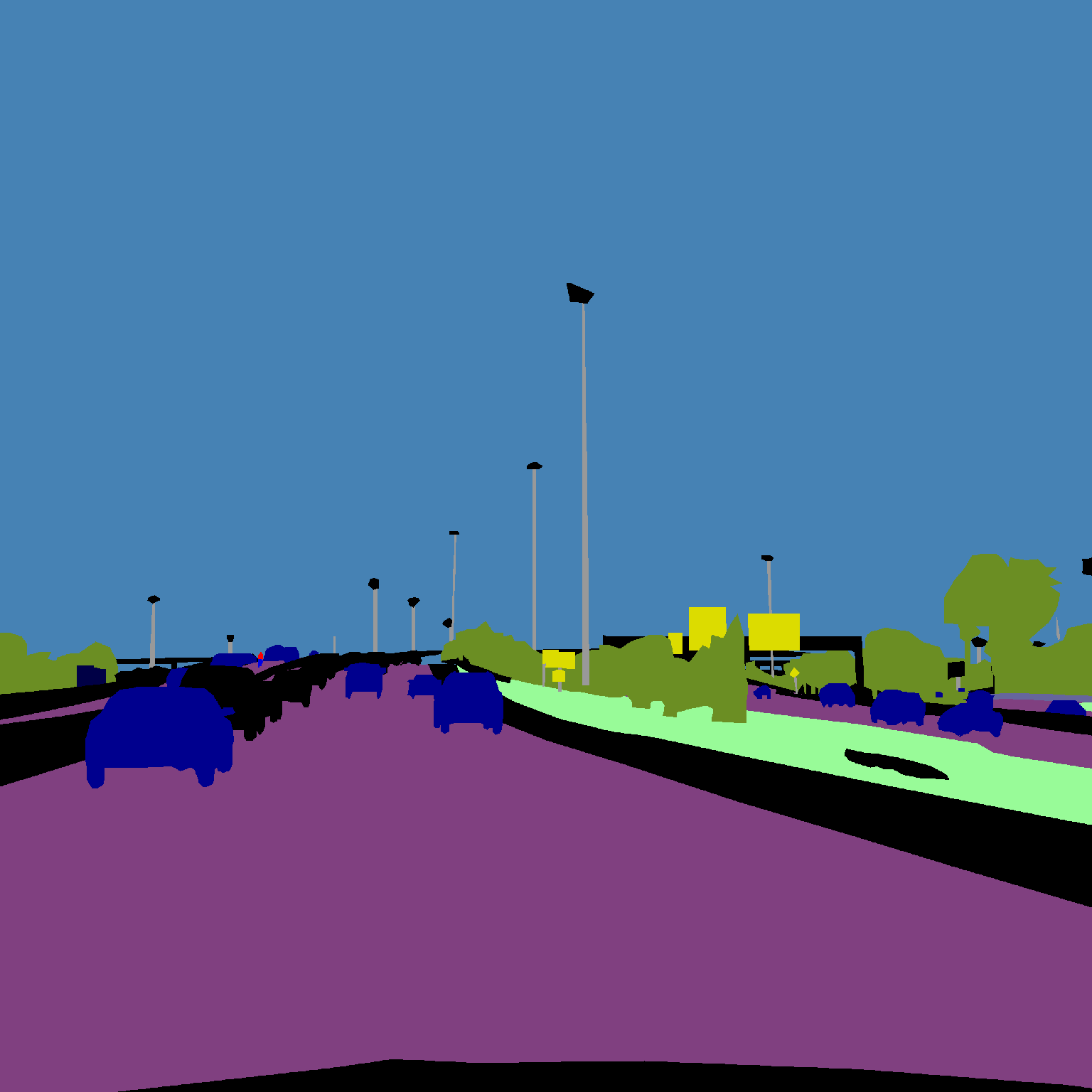} & 
		\includegraphics[width=0.16\linewidth,trim=0cm 6cm 0cm 15cm, clip]{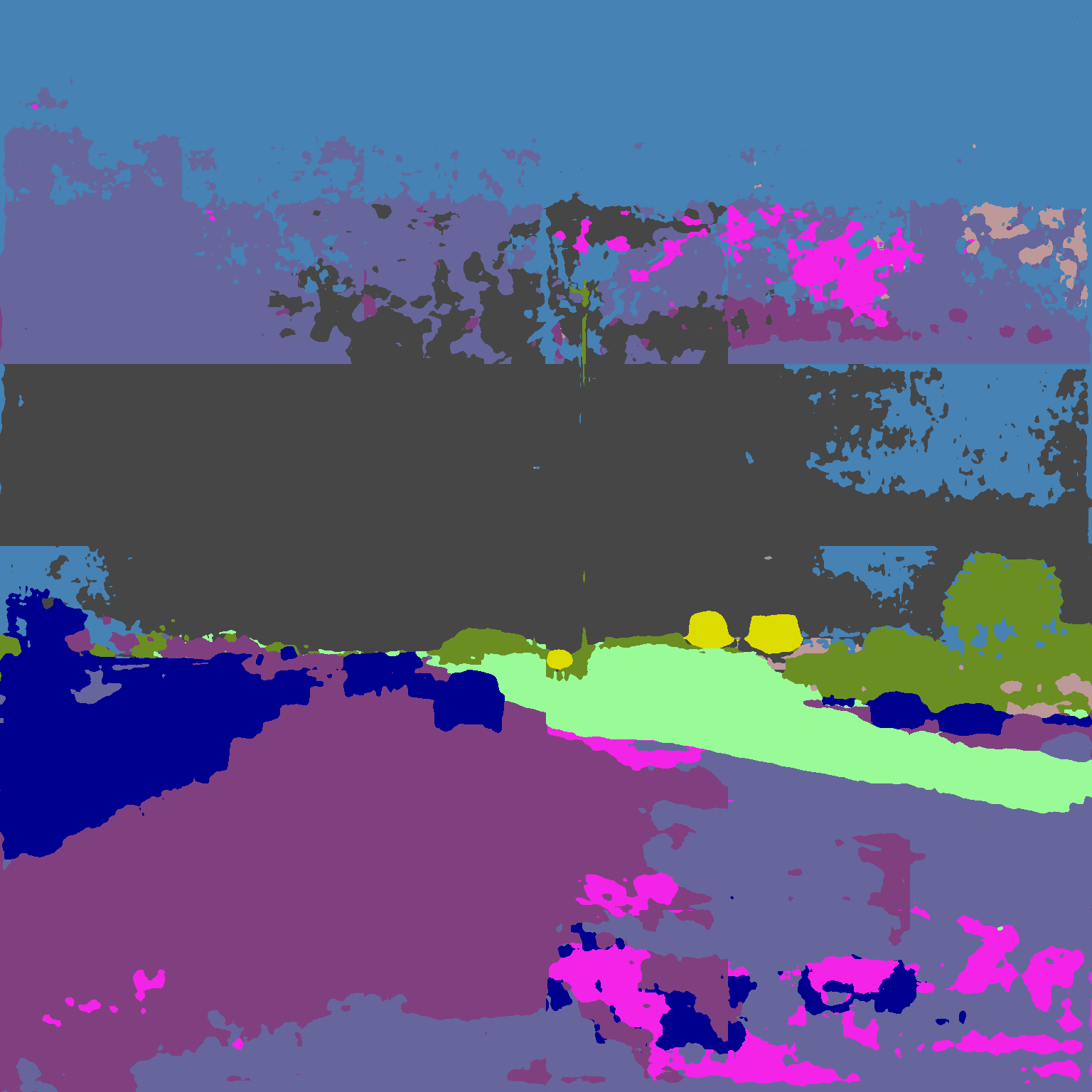} &
	    \includegraphics[width=0.16\linewidth,trim=0cm 6cm 0cm 15cm, clip]{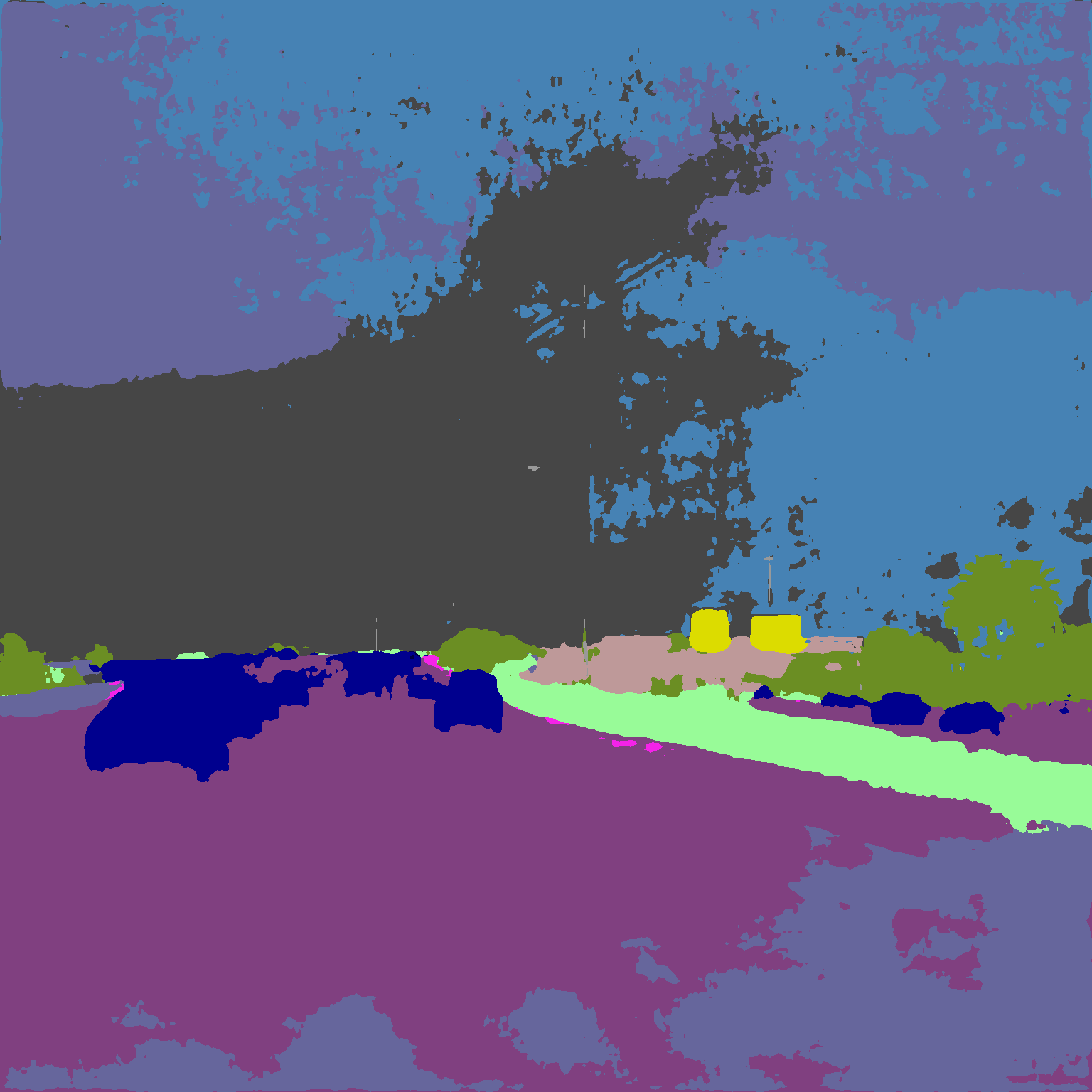}&
        \includegraphics[width=0.16\linewidth,trim=0cm 6cm 0cm 15cm, clip]{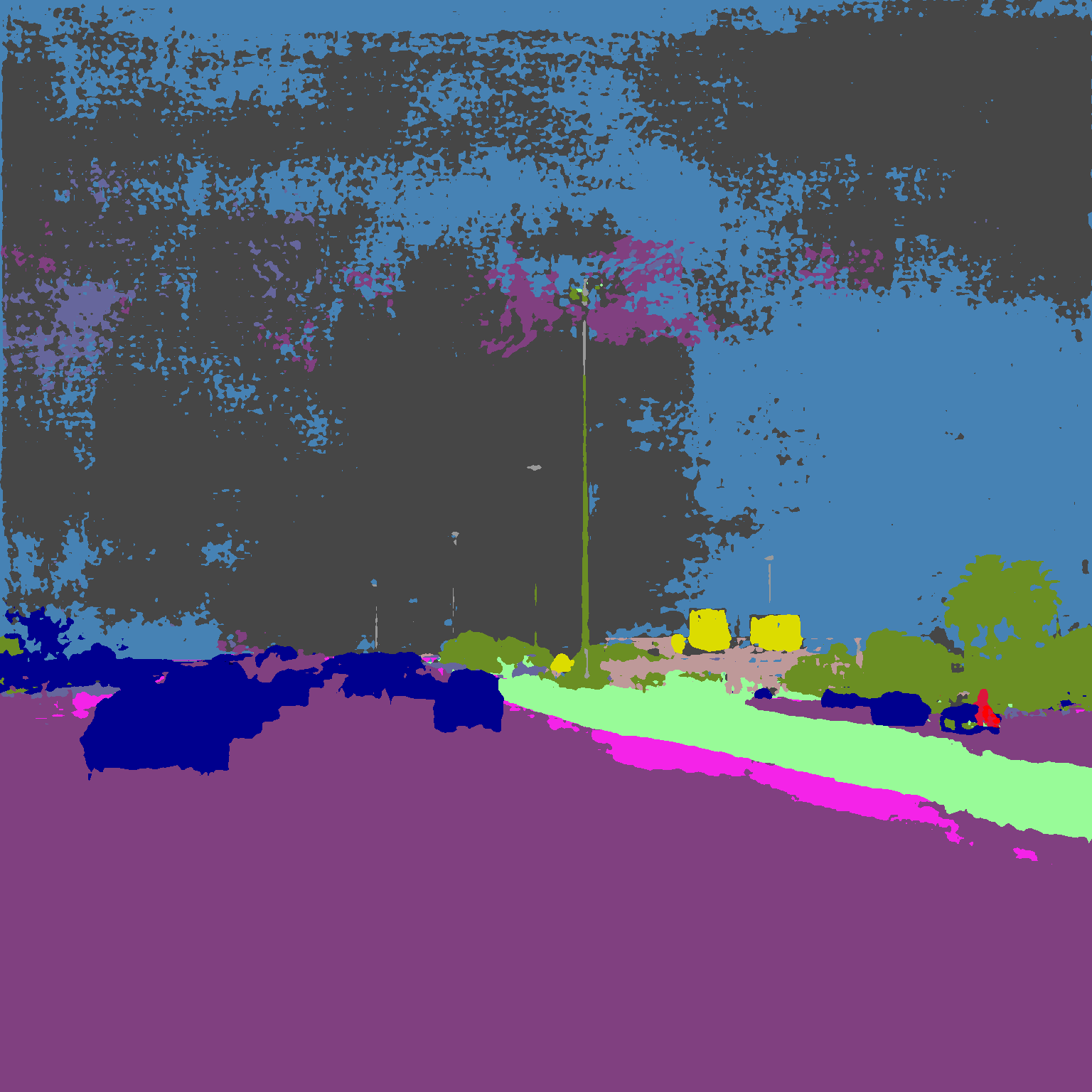}&
		\includegraphics[width=0.16\linewidth,trim=0cm 6cm 0cm 15cm, clip]{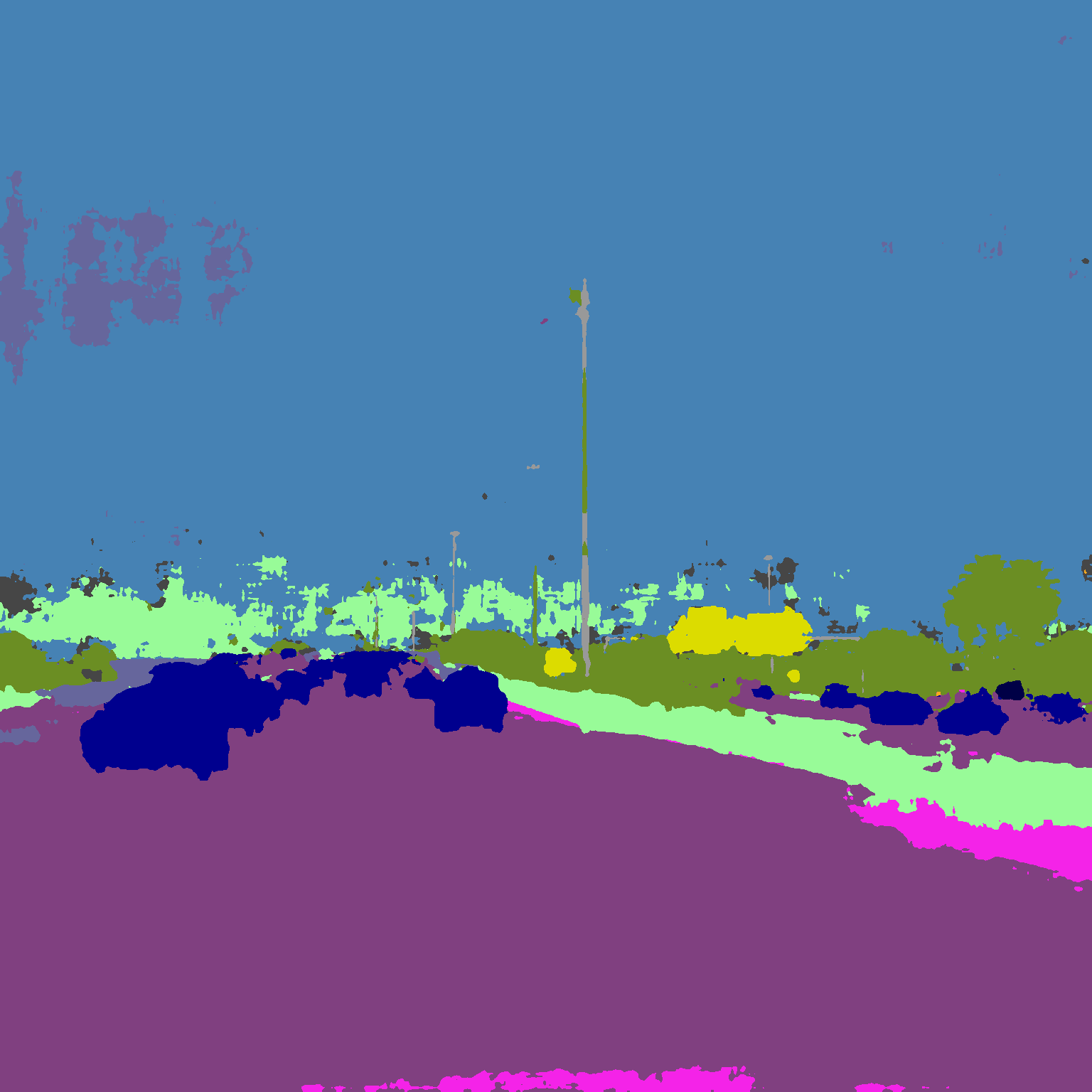} \\
		    \multirow{1}{*}[4em]{\texttt{S}} & \rowqual{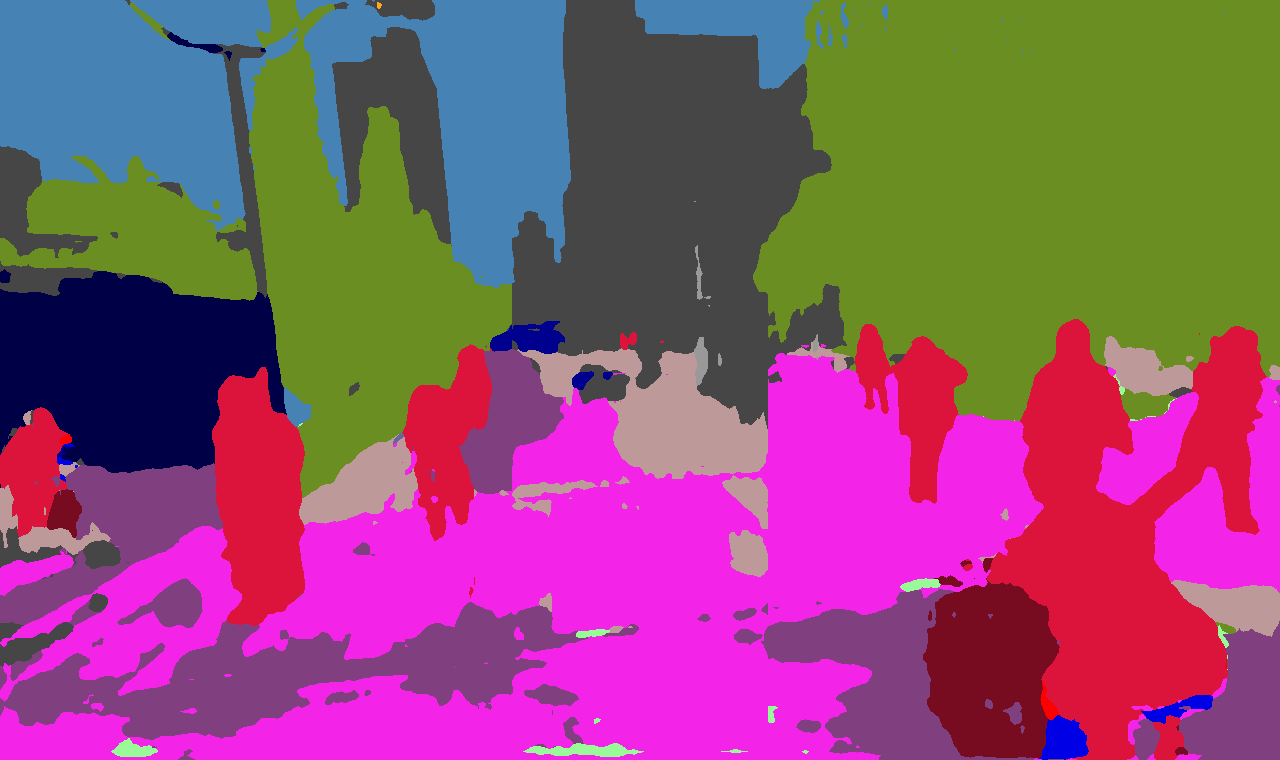}
	\end{tabular}
    \newcolumntype{C}[1]{>{\centering\let\newline\\\arraybackslash\hspace{0pt}}m{#1}}
    \begin{tabular}{C{5em}C{5em}C{5em}C{5em}C{5em}C{5em}C{5em}C{5em}C{5em}C{5em}}
    \tiny
    \cellcolor{csroad}{\textcolor{white}{Road}} & \cellcolor{csside}{\textcolor{black}{Sidewalk}} & \cellcolor{csbuild}{\textcolor{white}{Building}} & \cellcolor{cswall}{\textcolor{white}{Wall}} & \cellcolor{csfence}{\textcolor{black}{Fence}} & \cellcolor{cspole}{\textcolor{black}{Pole}} & \cellcolor{cslight}{\textcolor{black}{Traffic light}} & \cellcolor{cssign}{\textcolor{black}{Traffic sign}} & \cellcolor{csveg}{\textcolor{black}{Vegetation}} & \cellcolor{csterrain}{\textcolor{black}{Terrain}} \\
    \cellcolor{cssky}{\textcolor{white}{Sky}} & \cellcolor{csperson}{\textcolor{white}{Person}} & \cellcolor{csrider}{\textcolor{white}{Rider}} & \cellcolor{cscar}{\textcolor{white}{Car}} & \cellcolor{cstruck}{\textcolor{white}{Truck}} & \cellcolor{csbus}{\textcolor{white}{Bus}} & \cellcolor{cstrain}{\textcolor{white}{Train}} & \cellcolor{csmbike}{\textcolor{white}{Motorbike}} & \cellcolor{csbike}{\textcolor{white}{Bicycle}} & \cellcolor{csignore}{\textcolor{white}{n/a}} \\
    \end{tabular}
    \vspace{-0.1cm}
    \smallskip\caption{\textbf{Qualitative results.} \textit{Columns 1-2}: Image and ground truth (GT), \textit{Columns 3-4-5}: DGSS methods results, \textit{Column 6}: Our results. The models are trained on GTAV with \mbox{ResNet-50} backbone.}
    \vspace{-0.3cm}
    \label{fig:qualres}
\end{figure*}

\parag{Qualitative results.} We visually compare the segmentation results with Pin the memory~\cite{kim2022pin}, SHADE~\cite{zhao2022style} and WildNet~\cite{lee2022wildnet} in~\cref{fig:qualres}. \method clearly outperforms other DGSS methods on ``stuff'' (\eg,~road and sky) and ``things'' (\eg,~bicycle and bus) classes. 

\subsection{Decoder-Probing Fine-Tuning (DP-FT)}
\label{sec:dp_ft}

Kumar \textit{et al.}~\cite{kumar2022finetuning} show that standard fine-tuning may distort the pretrained feature representation, leading to degraded OOD performances for classification. Consequently, they propose a two-step training strategy: (1)~Training a linear probe (LP) on top of the frozen backbone features, (2)~Fine-tuning (FT) both the linear probe and the backbone. Inspired by it, Saito \textit{et al.}~\cite{saito2023mind} apply the same strategy for object detection, which is referred to as Decoder-probing Fine-tuning (DP-FT). They observe that DP-FT improves over DP depending on the architecture. We hypothesize that the effect is also dependent on the pretraining paradigm and the downstream task. As observed in~\cref{tab:imagenet_vs_CLIP}, CLIP might remarkably \textit{overfit} the source domain when fine-tuned. In~\cref{tab:dp_ft}, we compare fine-tuning (FT), decoder-probing (DP) and DP-FT. DP brings improvements over FT since it completely preserves the pretrained representation. Yet, DP major drawback lies in its limitation to adapt features for the downstream task, resulting in suboptimal results. Surprisingly, DP-FT largely falls behind DP, meaning that the 
learned features over-specialize to the source domain distribution
even with a ``decoder warm-up''.

The results advocate for the need of specific strategies to preserve CLIP robustness for semantic segmentation.  This need emerges from the additional gap between pretraining (\ie aligning object-level and language representations) and fine-tuning (\ie supervised pixel classification).

\begin{table}
	\setlength{\tabcolsep}{0.01\linewidth}
	\centering
  \resizebox{1.\linewidth}{!}{
	\begin{tabular}{c|cccccccc|c}
		\toprule
		  Method & \texttt{C} & \texttt{B} & \texttt{M} & \texttt{S} & \texttt{AN} & \texttt{AS} & \texttt{AR} & \texttt{AF} & Mean \\
		\midrule
		FT & 16.81 & 16.31 & 17.80 & 27.10 & 2.95 & 8.58 & 14.35 & 13.61 & 14.69\\
        DP & \underline{34.13} & \underline{37.67} & \underline{42.21} & 29.10 & \underline{10.71} & \underline{26.26} & \underline{29.47} & \underline{30.40} & \underline{29.99}\\
        DP-FT & 25.62 & 21.71 & 26.39 & \underline{31.45} & 4.22 & 18.26 & 20.07 & 20.85 & 21.07\\
        \rowcolor{ours} \method (ours) & \textbf{48.15} & \textbf{45.61} & \textbf{52.11} & \textbf{34.23} & \textbf{14.96} & \textbf{37.09} & \textbf{38.66} &  \textbf{40.25} & \textbf{38.88}\\
		\bottomrule
	\end{tabular}
  }
    \caption{
	\textbf{\method \vs DP-FT}. Performance (mIoU\%) of \method compared to Fine-tuning (FT), Decoder-probing (DP) and Decoder-probing Fine-tuning (DP-FT). We use here \mbox{ResNet-50}, trained on GTAV. We emphasize \textbf{best} and \underline{second best} results.
	}
\label{tab:dp_ft}
\end{table}

\begin{table} 
	\setlength{\tabcolsep}{0.01\linewidth}
	\centering
 \resizebox{1.\linewidth}{!}{
	\begin{tabular}{ccc|cccccccc|c}
		\toprule
		  Freeze & Augment & Mix & \texttt{C} & \texttt{B} & \texttt{M} & \texttt{S} & \texttt{AN} & \texttt{AS} & \texttt{AR} & \texttt{AF} & Mean\\
		\midrule
        \xmark & \xmark & \xmark & 16.81 & 16.31 & 17.80 & 27.10 & 2.95 & 8.58 & 14.35 & 13.61 & 14.69 \\
        \xmark & \checkmark & \xmark & 22.48 & 26.05 & 24.15 & 25.40 & 4.83 & 17.61 & 22.86 & 19.75 & 20.39\\
        \xmark & \xmark & \checkmark & 20.07 & 21.24 & 22.91 & 26.52 & 1.28 & 14.99 & 22.09 & 20.51 & 18.70 \\
        \xmark & \checkmark & \checkmark & 27.53 & 26.59 & 26.27 & 26.91 & 4.90 & 18.91 & 25.60 & 22.14 & 22.36\\
        \checkmark & \xmark & \xmark & 37.83 & 38.88 & 44.24 & 31.93 & 12.41 & 29.59 & 31.56 & 33.05 & 32.44\\
		  \checkmark & \checkmark & \xmark & 36.65 & 35.73 & 37.32 & 30.44 & \underline{14.72} & 34.65 & 34.91 & \underline{38.98} & 32.93\\
        \checkmark & \xmark & \checkmark & \underline{43.43} & \underline{43.79} & \underline{48.19} & \underline{33.70} & 11.32 & \underline{35.55} & \underline{36.15} & 38.19 & \underline{36.29} \\
	    \rowcolor{ours}\checkmark & \checkmark & \checkmark & \textbf{48.15} & \textbf{45.61} & \textbf{52.11} & \textbf{34.23} & \textbf{14.96} & \textbf{37.09} & \textbf{38.66} &  \textbf{40.25} & \textbf{38.88} \\
    \bottomrule
	\end{tabular}
}
    \caption{
	\textbf{Ablation of \method components.} Performance (mIoU~\%) after removing one or more components of \method.
	}
\label{tab:FAMIX_ing_ablation}
\end{table}

\subsection{Ablation studies}
\label{sec:ablate}
We conduct all the ablations on a \mbox{ResNet-50} backbone with GTAV (\texttt{G}) as source dataset.

\parag{Removing ingredients from the recipe.} \method is based on minimal fine-tuning of the backbone (\ie, Freeze), style augmentation and mixing. We show in~\cref{tab:FAMIX_ing_ablation} that the best generalization results are only obtained when combining the \textit{three} ingredients. Specifically, when the backbone is fine-tuned (\ie, Freeze \xmark), the performances are largely harmed. When minimal fine-tuning is performed (\ie, Freeze \checkmark), we argue that the augmentations are too strong to be applied without style mixing; the latter brings both effects of domain interpolation and use of the original statistics. Subsequently, when style mixing is not applied (\ie Freeze \checkmark, Augment \checkmark, Mix \xmark), the use of mined styles brings mostly no improvement on OOD segmentation compared to training without augmentation (\ie Freeze \checkmark, Augment \xmark, Mix \xmark). 
Note that for Freeze \checkmark, Augment \checkmark, Mix \xmark, the line 8 in~\cref{algo:training_FAMIX} becomes: 
\begin{equation}
    \src{\mb{f}}^{(ij)} \gets \adain(\src{\mb{f}}^{(ij)},\bs{\mu}^{(ij)}, \bs{\sigma}^{(ij)}) 
\end{equation}
Our style mixing is different from MixStyle~\cite{zhou2021domain} for being applied: (1) patch-wise and (2) between original styles of the source data and augmented versions of them. Note that the case (Freeze \checkmark, Augment \xmark, Mix \checkmark) could be seen as a variant of MixStyle, yet applied locally and class-wise. Our complete recipe is proved to be significantly more effective with a boost of $\approx+6$ mean mIoU w.r.t. the baseline of training without augmentation and mixing.

\parag{Prompt construction.} ~\cref{tab:class_name_ablation} reports results when ablating the prompt construction.
In FAMix, the final prompt is derived by concatenating \texttt{$<$random\! style\! prompt$>$} and \texttt{$<$class\! name$>$}; removing either of those leads to inferior results.
Interestingly, replacing the style prompt by random characters -- \eg ``ioscjspa'' -- does not significantly degrade the performance.
In certain aspects, using random prompts still induces a randomization effect within the \method framework.
However, meaningful prompts still consistently lead to the best results.

\begin{table}[t]
	\setlength{\tabcolsep}{0.01\linewidth}
	\centering
	\resizebox{1.0\linewidth}{!}{%
		\begin{tabular}{ccc|cccccccc|c}
			\toprule
		 \makecell{RCP} & \makecell{RSP} & \makecell{CN} & \texttt{C} & \texttt{B} & \texttt{M} & \texttt{S} & \texttt{AN} & \texttt{AS} & \texttt{AR} & \texttt{AF} & Mean \\
			\midrule
		     & & \checkmark & 45.99 & 43.71 & \underline{50.48} & \textbf{34.75} & \underline{15.22} & 35.09 & 34.92 & 38.17 & 37.29 \\
		 \checkmark & & & 46.10 & 44.24 & 48.90 & 33.62 & 13.39 &  35.99 & 36.68 & \underline{39.86} & 37.35 \\
		    & \checkmark & & 45.64	& 44.59	& 49.13	& 33.64	& \textbf{15.33} & \textbf{37.32} & 35.98 & 38.85 & 37.56 \\
		    \checkmark & & \checkmark & \underline{47.83} & \underline{44.83} & 50.38 & \underline{34.27} & 14.43 & 37.07 & \underline{37.07} & 38.76 & \underline{38.08} \\
			\rowcolor{ours} & \checkmark & \checkmark & \textbf{48.15} & \textbf{45.61} & \textbf{52.11} & 34.23 & 14.96 & \underline{37.09} & \textbf{38.66} &  \textbf{40.25} &\textbf{ 38.88} \\
			\bottomrule
		\end{tabular}%
	}
 \vspace{-0.1cm}
	\caption{\textbf{Ablation on the prompt construction.} Performance (mIoU \%) for different prompt constructions. RCP, RSP and CN refer to $<$\texttt{random\! character\! prompt}$>$, $<$\texttt{random\! style\! prompt}$>$ and $<$\texttt{class\! name}$>$, respectively.}
 \vspace{-0.3cm}
\label{tab:class_name_ablation}
\end{table}

\begin{figure}
        \centering
	\begin{subfigure}{0.458\linewidth}
		\includegraphics[width=1.0\linewidth]{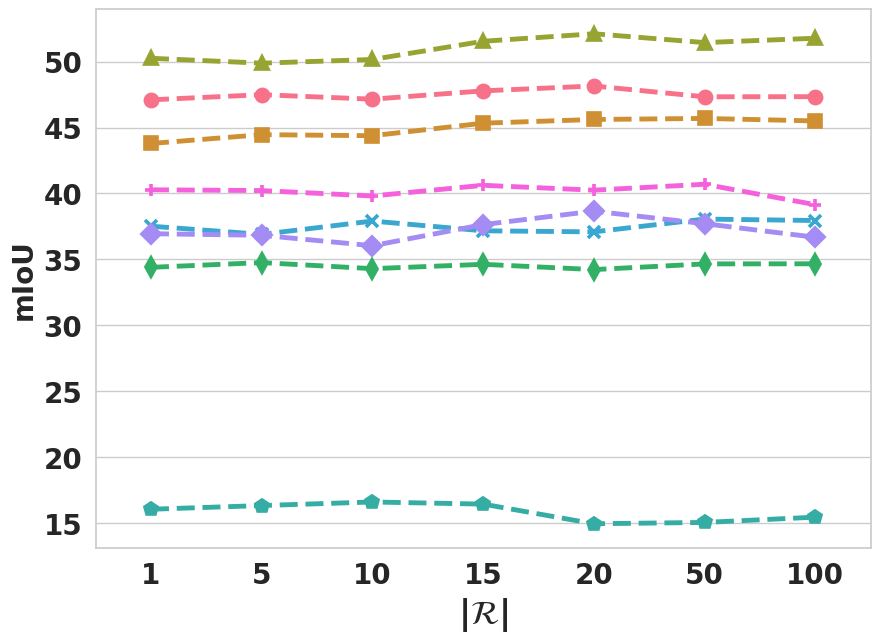}
		\caption{Number of prompts}
		\label{fig:ablate_card_R}
	\end{subfigure}\hfill%
	\begin{subfigure}{0.541\linewidth}
		\includegraphics[width=1.0\linewidth]{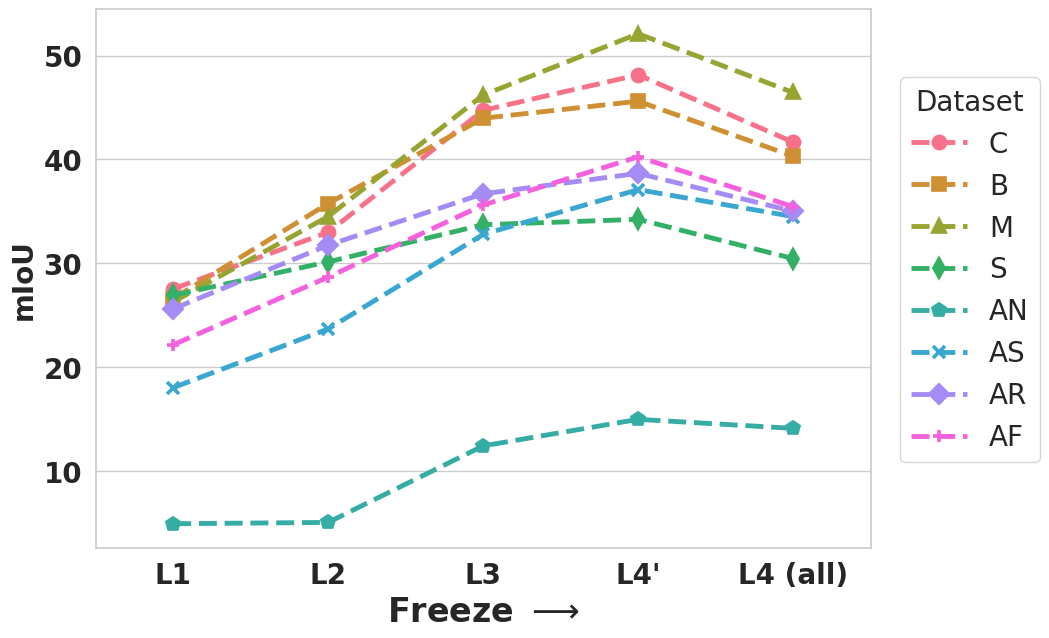}
		\caption{Effect of layer freezing}
		\label{fig:ablate_freeze}
	\end{subfigure}
    \vspace{-0.3cm}
    \caption{\textbf{Ablation of prompt set and freezing strategy.}
    \subref{fig:ablate_card_R}~Performance (mIoU \%) on test datasets w.r.t. the number of random style prompts in $\mc{R}$.
    \subref{fig:ablate_freeze}~Effect of freezing layers reporting on x-axis the last frozen layer. For example, `L3' means freezing L1, L2 and L3. `L4'\,' indicates that the \textit{Layer4} is partially frozen.}
    \vspace{-0.3cm}
\end{figure}

\parag{Number of style prompts.} 
\method uses a set \mc{R} of random style prompts which are concatenated with the class names; \mc{R} is formed by querying ChatGPT\footnote{\url{https://chat.openai.com/}} using \texttt{$<$give\! me\! 20\! prompts\! of\! 2\! to\! 5\! words\! describing\! random\! image\! styles$>$}. The output prompts are provided in~\cref{sec:prompts_used}. 
\cref{fig:ablate_card_R} shows that the 
size of $\mc{R}$ has a marginal impact on \method performance.
Yet, the mIoU scores on \texttt{C}, \texttt{B}, \texttt{M} and \texttt{AR} are higher for $|\mc{R}|=20$ compared to $|\mc{R}|=1$ and almost equal for the other datasets.

The low sensitivity of the performance to the size of $\mc{R}$ could be explained by two factors. First, mining even from a single prompt results in different style variations as the optimization starts from different anchor points in the latent space, as argued in~\cite{fahes2023poda}. Second, mixing style between the \textit{source} and the mined \textit{proxy} domains is the crucial factor making the network explore intermediate domains during training.
This does not contradict the effect of our prompt construction which leads to the best results (\cref{tab:class_name_ablation}).

\parag{Local \vs global style mining.} To highlight the effect of our class-wise local style mining, we perform an ablation replacing it with global style mining. Specifically, the same set of \texttt{$<$random\! style\! prompt$>$} are used, though being concatenated with \texttt{$<$driving$>$} as a global description instead of local class name. Intuitively, local style mining and mixing induces richer style variations and more contrast among patches. The results in~\cref{tab:global_vs_local} show the effectiveness of our local style mining and mixing strategy, bringing about $3$ mIoU improvement on \mbox{\texttt{G} $\rightarrow$ \texttt{C}}. 

\begin{table}[t]
	\setlength{\tabcolsep}{0.01\linewidth}
	\centering
 \newcolumntype{H}{>{\setbox0=\hbox\bgroup}c<{\egroup}@{}}
 \resizebox{1.\linewidth}{!}{
	\begin{tabular}{ccH|cccccccc|c}
		\toprule
		  \multicolumn{2}{c}{Syle mining} & & \texttt{C} & \texttt{B} & \texttt{M} & \texttt{S} & \texttt{AN} & \texttt{AS} & \texttt{AR} & \texttt{AF} & Mean \\
		\midrule
       \multirow{5}{*}{global w/} & ``street view'' & & 45.51 & 45.12 & 50.40 & 33.65 & 14.59 & 36.92 & 37.38 & 40.53 & 38.01\\
        & ``urban scene'' & & 46.59 & 45.38 & 51.33 & 33.67 & 14.42 & 35.96 & 37.30 & 40.52 & 38.15 \\
        & ``roadscape'' & & 45.49 & 45.55 & 50.63 & 33.66 & 14.77 & 36.75 & 37.07 & 40.33 & 38.03\\
        & ``commute snapshot'' & & 45.39 & 45.08 & 50.50 & 33.68 & 13.65 & 36.63 & 37.93 & \textbf{40.92} & 37.97\\
    	& ``driving''  & & 45.06 & 44.98 & 50.67 & 33.36 & 14.84 & 35.11 & 36.21 & 39.52 & 37.47\\
      
    	\rowcolor{ours} local & & & \textbf{48.15} & \textbf{45.61} & \textbf{52.11} & \textbf{34.23} & \textbf{14.96} & \textbf{37.09} & \textbf{38.66} &  40.25 & \textbf{38.88} \\
		\bottomrule
	\end{tabular}
 }
  \caption{\textbf{Ablation on style mining.} Global style mining consists of mining one style per feature map, using $<$\texttt{random\! style\! prompt}$>$ + $<$\texttt{global\! class}$>$ as prompt.}
	
\label{tab:global_vs_local}
\end{table}

\parag{What to mix?} Let $\mc{S}=\bigcup_{k=1}^{K} \mc{S}^{(k)}$
and $\mc{T}= \bigcup_{k=1}^{K} \mc{T}^{(k)}$ the sets of class-wise source and augmented features, respectively.  In \method training, for an arbitrary patch $\src{\mb{f}}^{(ij)}$, style mixing is performed between the original source statistics and statistics sampled from the augmented set (\ie, $(\bs{\mu}^{(ij)}, \bs{\sigma}^{(ij)}) \in \mc{T}^{(c_p^{(ij)})}$, see~\eqref{eq:mu_mix} and~\eqref{eq:sigma_mix}). 
In class-wise vanilla MixStyle, $(\bs{\mu}^{(ij)}, \bs{\sigma}^{(ij)}) \in \mc{S}^{(c_p^{(ij)})}$. In~\cref{tab:what_to_mix}, we show that sampling $(\bs{\mu}^{(ij)}, \bs{\sigma}^{(ij)})$ from $\mc{S}^{(c_p^{(ij)})} \cup \mc{T}^{(c_p^{(ij)})}$ does not lead to better generalization, despite sampling from a set with twice the cardinality. This supports our mixing strategy visualized in~\cref{fig:aug_mix}. Intuitively, sampling from $\mc{S} \cup \mc{T}$ could be viewed as applying either MixStyle or our mixing with a probability $p=0.5$.\\
\begin{table}[t]

	\setlength{\tabcolsep}{0.01\linewidth}
	\centering
 \resizebox{1.\linewidth}{!}{
	\begin{tabular}{c|cccccccc|c}
		\toprule
		  Syle mining & \texttt{C} & \texttt{B} & \texttt{M} & \texttt{S} & \texttt{AN} & \texttt{AS} & \texttt{AR} & \texttt{AF} & Mean \\
		\midrule
		  $\mc{S}$  & 43.43 & 43.79 &  48.19 & 33.70 & 11.32 & 35.55 & 36.15 & 38.19 & 36.29   \\
        $\mc{S} \cup \mc{T}$ & 44.76 & 45.59 & 50.78 &  34.05 & 13.67 & 36.92 & 37.18 & 38.13 & 37.64  \\
		\rowcolor{ours}$\mc{T}$ (ours) & \textbf{48.15} & \textbf{45.61} & \textbf{52.11} & \textbf{34.23} & \textbf{14.96} & \textbf{37.09} & \textbf{38.66} & \textbf{40.25} & \textbf{38.88}  \\ 
        
		\bottomrule
	\end{tabular}
 }
 \caption{\textbf{Ablation on the sets used for mixing.} The styles $(\bs{\mu}, \bs{\sigma})$ used in~\eqref{eq:mu_mix} and~\eqref{eq:sigma_mix} are sampled either from $\mc{S}$ or $\mc{S} \cup \mc{T}$ or $\mc{T}$.} 
	\label{tab:what_to_mix}
\end{table}

\parag{Minimal fine-tuning.} We argue for minimal fine-tuning as a compromise between pretrained feature preservation and adaptation. ~\cref{fig:ablate_freeze} shows an increasing OOD generalization trend with more freezing. Interestingly, only fine-tuning the last layers of the last convolutional block (where the dilation is applied) achieves the best results. When training on Cityscapes (\cref{tab:cstrain}), we observed that freezing all the layers except \textit{Layer4} achieves the best results.

\subsection{Does \method require language?}
\label{sec:language_vs_noise}

Inspired by the observation that target statistics deviate around the source ones in real cases~\cite{fan2023towards}, we conduct an experiment where we replace language-driven style mining by noise perturbation. The same procedure of \method is kept: (i) Features are divided into patches, perturbed with noise and then saved into a style bank based on the dominant class; (ii) During training, patch-wise style mixing of original and perturbed styles is performed.

Different from Fan \etal~\cite{fan2023towards}, who perform a perturbation on the feature statistics using a normal distribution with pre-defined parameters, we experiment perturbation with different magnitudes of noise controlled by the signal-to-noise ratio (SNR). Consider the mean of a patch $\mu \in \mathbb{R}^c$ as a signal, the goal is to perturb it with some noise $n_\mu \in \mathbb{R}^{c}$. The $\texttt{SNR}_{\texttt{dB}}$ between $\|\mu\|$ and $\|n_\mu\|$ is defined as $\texttt{SNR}_{\texttt{dB}}= 20 \, \text{log}_{10} \left( \nicefrac{\|\mu\|}{\|n_\mu\|}\right)$. 
Given $\mu$, $\texttt{SNR}_{\texttt{dB}}$, and  $n \sim \mathcal{N}(0, I)$, where $I \in \mathbb{R}^{c \times c}$ is the identity matrix, the noise is computed as $n_{\mu} = 10 ^ {\frac{-\texttt{SNR}}{20}} \frac{\|\mu\|}{\|n\|}n$.
We add $\mu+n_\mu$ to the style bank corresponding to the dominant class in the patch. The same applies to $\sigma \in \mathbb{R}^{c}$. The results of training for different noise levels are in~\cref{tab:prompt_vs_noise}. Using language as source of randomization outperforms any noise level. The baseline corresponds to the case where no augmentation nor mixing are performed (See~\cref{tab:FAMIX_ing_ablation}, Freeze \checkmark, Augment \xmark, Mix \xmark). SNR=$\infty$ could be seen as a variant of MixStyle, applied class-wise to patches (See~\cref{tab:FAMIX_ing_ablation}, Freeze \checkmark, Augment \xmark, Mix \checkmark). The vanilla MixStyle gets inferior results.

Besides lower OOD performance, one more disadvantage of noise augmentation compared to our language-driven augmentation is the need to select a value for the SNR, for which the optimal value might vary depending on the target domain encountered at the test time.
\begin{table}[t]

	\setlength{\tabcolsep}{0.01\linewidth}
	\centering
    \resizebox{1.\linewidth}{!}{
	\begin{tabular}{c|cccccccc|c}
		\toprule
		  SNR & \texttt{C} & \texttt{B} & \texttt{M} & \texttt{S} & \texttt{AN} &  \texttt{AS} &  \texttt{AR} & \texttt{AF} & Mean\\
		\midrule
  		  Baseline & 37.83 & 38.88 & 44.24 & 31.93 & 12.41 & 29.59 & 31.56 & 33.05 & 32.44 \\
        \midrule
		  5 & 28.78 & 29.24 & 30.32 & 21.67 & 12.60 & 24.00 & 25.95 & 25.87 & 24.80\\
		10 & 40.09 & 39.50 & 43.45 & 29.09 & 13.36 & 33.47 & 33.11 & 36.17 & 33.53\\
        15 & 45.02 & 44.16 & 48.63 & 32.96 & \underline{14.55} & \underline{36.09} & 35.99 & \textbf{40.96} & \underline{37.30}\\
        20 & \underline{45.52} & \underline{44.29} & \underline{49.26} & 33.45  & 12.40 & 35.96 & \underline{36.52} & 38.60 & 37.00\\
        25 & 44.82 & 44.26 & 48.54 & 33.30 & 11.38 & 34.51 & 35.46 & 37.61 &  36.24\\
        30 & 43.07 & 43.80 & 48.31 & 33.47 & 12.33 & 35.05 & 35.58 & 38.10 & 36.21\\
        $\infty$ & 43.43 & 43.79 & 48.19 & \underline{33.70} & 11.32 & 35.55 & 36.15 & 38.19 & 36.29 \\
        \midrule
        MixStyle~\cite{zhou2021domain} & 40.97 & 42.04 & 48.36 & 33.15 & 13.14 & 31.26 & 34.94 & 38.12 & 35.25 \\
        \midrule
        \rowcolor{ours} Prompts 
        & \textbf{48.15} & \textbf{45.61} & \textbf{52.11} & \textbf{34.23} & \textbf{14.96} & \textbf{37.09} & \textbf{38.66} & \underline{40.25} & \textbf{38.88}  \\
  \bottomrule
	\end{tabular}
  }
        \smallskip\caption{\textbf{Noise vs prompt-driven augmentation.} 
        The prompt-driven augmentation in FAMix is replaced by random noise with different levels defined by SNR.
        We also include vanilla MixStyle. 
        The prompt-driven strategy is superior.
	}
\label{tab:prompt_vs_noise}
\end{table}

\section{Conclusion}
\label{sec:conclusion}

We presented \method, a simple recipe for domain generalized semantic segmentation with CLIP pretraining. We proposed to locally mix the styles of source features with their augmented counterparts obtained using language prompts. Combined with minimal fine-tuning, \method significantly outperforms the state-of-the-art approaches. Extensive experiments showcase the effectiveness of our framework. We hope that \method will serve as a strong baseline in future works, exploring the potential of leveraging large-scale vision-language models for perception tasks.

\noindent\textbf{Acknowledgment.} This work was partially funded by French project SIGHT (ANR-20-CE23-0016) and was supported by ELSA - European Lighthouse on Secure and Safe AI funded by the European Union under grant agreement No.~101070617. It was performed using HPC resources from GENCI–IDRIS (Grant AD011014477).
\newcommand{\var}[1]{{\small\,$\pm$#1}}
\newcommand{\vartn}[1]{{\scriptsize\,$\pm$#1}}

\section*{Appendix}
\appendix
\renewcommand{\appendixautorefname}{Section}

\
We provide details about~\cref{tab:imagenet_vs_CLIP} experiments in~\cref{sec:clip_vs_imgnet}. Further, we report class-wise performance for \method in~\cref{sec:class_perf}, and detail the prompts used in our experiments in~\cref{sec:prompts_used}. Finally, we
discuss the limitations and perspectives in~\cref{sec:limit}.

We refer to the \textbf{supplementary video} for further demonstration of FAMix qualitative performance: \mbox{\url{https://youtu.be/vyjtvx2El9Q}}.

\section{CLIP \vs ImageNet initialization}
\label{sec:clip_vs_imgnet}
In~\cref{tab:imagenet_vs_CLIP} of the main paper, we introduce a comparison of ImageNet and CLIP pretraining for out-of-distribution semantic segmentation. We clarify here its implementation.

To produce~\cref{tab:imagenet_vs_CLIP} we employ the public code\footnote{\url{https://github.com/VainF/DeepLabV3Plus-Pytorch}} and finetuned with SGD using a learning rate of $10^{-1}$ for the segmenter and $10^{-2}$ for the backbone.
We note that we freeze the \textit{stem layers} and \textit{Layer1} for both backbones, \ie, ImageNet and CLIP initialized ResNet-50, after observing that full fine-tuning leads to subpar in-domain results for CLIP. Crucially, 
in this setting, \textit{both} ImageNet and CLIP initialized networks converge and achieve the same performance in-domain (on GTA5 validation set). Hence, we argue that the poor OOD performance of CLIP initialization in~\cref{tab:imagenet_vs_CLIP} may originate from the distortion of the robust CLIP representation towards the source domain distribution as advocated in~\cite{kumar2022finetuning}. 
To alleviate such distortion, we freeze most of the backbone layers in \method{}.

However, we highlight that different hyper-parameter choices could boost the performance to some extent. For example, Rao \textit{et al.}~\cite{rao2022denseclip} observed that fine-tuning CLIP for semantic segmentation with the default configuration in MMSegmentation\footnote{\url{https://github.com/open-mmlab/mmsegmentation}} leads to 15.6\% mIoU lower performance than its ImageNet pre-trained counterpart on ADE20K~\cite{zhou2019semantic}. Consequently, they propose using AdamW~\cite{loshchilov2018decoupled} for optimization.

In \cref{tab:adamw_sgd} we reproduce the experiment of~\cref{tab:imagenet_vs_CLIP} experiment but using AdamW as optimizer.
$O_1$ refers to the optimization configuration adopted in our paper, \ie, SGD optimizer with a learning rate of $10^{-1}$ for the segmenter and $10^{-2}$ for the backbone. $O_2$ and $O_3$ both refer to the use of AdamW with a learning rate of $10^{-4}$ for the segmenter and $10^{-5}$ for the backbone. In $O_2$ all the backbone is fine-tuned while in $O_3$ the \textit{stem layers} and \textit{Layer1} are frozen, similar to $O_1$. Results show that using AdamW improves performance across out-of-distributions (OOD) domains for both CLIP and ImageNet initialized networks, but still largely lags behind \method (mean mIoU=38.88\%) and even our variant (Freeze \checkmark, Augment \xmark, Mix \xmark) (mean mIoU=32.44\%) in~\cref{tab:FAMIX_ing_ablation}.

These results hint that using AdamW with relatively low learning rates might reduce the feature distortion of CLIP. Motivated by this observation, one could question the necessity of the minimal fine-tuning part of \method, and whether similar results could be achieved only by augmenting, mixing, and fine-tuning with AdamW and low learning rate. We call this variant AMix (Augment and Mix) and show the results in~\cref{tab:amix_vs_famix}, which support the necessity of our full recipe.

\begin{table}[t]
	\setlength{\tabcolsep}{0.01\linewidth}
	\centering
 \resizebox{1.\linewidth}{!}{
	\begin{tabular}{cc|cccccccc|c}
		\toprule
		   Optim. & Pretraining & \texttt{C} & \texttt{B} & \texttt{M} & \texttt{S} & \texttt{AN} & \texttt{AS} & \texttt{AR} & \texttt{AF} & Mean \\
		\midrule
            \multirow{2}{*}{$O_1$} & ImageNet & \textbf{29.04} & \textbf{32.17} & \textbf{34.26} & \textbf{29.87} & \textbf{4.36} & \textbf{22.38} & \textbf{28.34} & \textbf{26.76} & \textbf{25.90}\\
            & CLIP & 16.81 & 16.31 & 17.80 & 27.10 & 2.95 & 8.58 & 14.35 & 13.61 & 14.69 \\
            \midrule
            \multirow{2}{*}{$O_2$} & ImageNet & 28.00 & \textbf{36.82} & \textbf{37.00} & 30.60 & \textbf{3.56} & \textbf{24.14} & \textbf{29.51} & \textbf{26.23} & \textbf{26.98}\\
            & CLIP & \textbf{31.73} & 25.89 & 30.68 & \textbf{33.32} & 2.56 & 19.17 & 21.42 & 17.58 & 22.79 \\
            \midrule
            \multirow{2}{*}{$O_3$} & ImageNet & \textbf{28.74} & \textbf{36.91} & \textbf{37.86} & 30.32 & \textbf{4.46} & \textbf{22.48} & \textbf{28.49} & \textbf{25.38} & \textbf{26.83}\\
            & CLIP & 26.81 & 23.11 & 29.82 & \textbf{32.38} & 4.20 & 18.50 & 22.59 & 20.31 & 22.22\\ 
            \bottomrule
	\end{tabular}
 }
  \caption{\textbf{Effect of optimization configurations on OOD performance.} Performance (mIoU \%) of CLIP \vs ImageNet initialized networks for different optimization configurations.}
	
\label{tab:adamw_sgd}
\end{table}

\begin{table}[!h]
	\setlength{\tabcolsep}{0.01\linewidth}
	\centering
 \resizebox{1.\linewidth}{!}{
	\begin{tabular}{c|cccccccc|c}
		\toprule
		  Method & \texttt{C} & \texttt{B} & \texttt{M} & \texttt{S} & \texttt{AN} & \texttt{AS} & \texttt{AR} & \texttt{AF} & Mean \\
		\midrule
            AMix & 40.50 & 38.69 & 36.05 & 33.61 & 4.03 & 23.03 & 30.01 & 26.89 & 29.10\\
            \method{} & \textbf{48.15} & \textbf{45.61} & \textbf{52.11} & \textbf{34.23} & \textbf{14.96} & \textbf{37.09} & \textbf{38.66} & \textbf{40.25} & \textbf{38.88} \\
            \bottomrule
	\end{tabular}
 }
  \caption{\textbf{AMix with AdamW optimizer \vs \method.} Performance (mIoU \%) of \method in our default configuration compared to a variant with no minimal fine-tuning, replacing SGD with AdamW optimizer.}
	
\label{tab:amix_vs_famix}
\end{table}

\section{Class-wise performance}
\label{sec:class_perf}

\noindent We report class-wise IoUs in~\cref{tab:gta5_rn50} and~\cref{tab:gta5_rn101}. The standard deviations of the mIoU (\%) over three runs are also reported.

\begin{table*}[t!]
	\setlength{\tabcolsep}{0.007\linewidth}
        \renewcommand{\arraystretch}{1.2}
        \newcommand{\best}{\bf}
		\centering
		\resizebox{\linewidth}{!}{
	  \begin{tabular}{c c c c c c c c c c c c c c c c c c c c | c}
			\toprule
			 \makecell{Target\\eval.}
			& \rotatebox{90}{\textcolor{csroad}{\rule{0.5em}{1.5ex}} road}
			& \rotatebox{90}{\textcolor{csside}{\rule{0.5em}{1.5ex}} sidewalk}
			& \rotatebox{90}{\textcolor{csbuild}{\rule{0.5em}{1.5ex}} building} 
			& \rotatebox{90}{\textcolor{cswall}{\rule{0.5em}{1.5ex}} wall} 
			& \rotatebox{90}{\textcolor{csfence}{\rule{0.5em}{1.5ex}} fence} 
			& \rotatebox{90}{\textcolor{cspole}{\rule{0.5em}{1.5ex}} pole} 
			& \rotatebox{90}{\textcolor{cslight}{\rule{0.5em}{1.5ex}} traffic light} 
			& \rotatebox{90}{\textcolor{cssign}{\rule{0.5em}{1.5ex}} traffic sign} 
			& \rotatebox{90}{\textcolor{csveg}{\rule{0.5em}{1.5ex}} vegetation} 
			& \rotatebox{90}{\textcolor{csterrain}{\rule{0.5em}{1.5ex}} terrain} 
			& \rotatebox{90}{\textcolor{cssky}{\rule{0.5em}{1.5ex}} sky} 
			& \rotatebox{90}{\textcolor{csperson}{\rule{0.5em}{1.5ex}} person} 
			& \rotatebox{90}{\textcolor{csrider}{\rule{0.5em}{1.5ex}} rider} 
			& \rotatebox{90}{\textcolor{cscar}{\rule{0.5em}{1.5ex}} car} 
			& \rotatebox{90}{\textcolor{cstruck}{\rule{0.5em}{1.5ex}} truck} 
			& \rotatebox{90}{\textcolor{csbus}{\rule{0.5em}{1.5ex}} bus} 
			& \rotatebox{90}{\textcolor{cstrain}{\rule{0.5em}{1.5ex}} train} 
			& \rotatebox{90}{\textcolor{csmbike}{\rule{0.5em}{1.5ex}} motorcycle} 
			& \rotatebox{90}{\textcolor{csbike}{\rule{0.5em}{1.5ex}} bicycle} 
			& mIoU\%\\
			\midrule

    \texttt{C} & 88.97 & 39.48 & 83.12 & 28.52 & 29.06 & 38.64 & 42.67 & 36.26 & 86.48 & 24.70 & 78.26 & 69.08 & 23.88 & 85.35 & 29.63 & 38.82 & 9.28 & 38.02 & 44.68 & 48.15\vartn{0.38}\\
    \cmidrule{2-21}
    \texttt{B} & 87.83 & 40.33 & 78.44 & 15.88 & 35.20 & 38.13 & 40.63 & 29.49 & 77.52 & 31.19 & 90.80 & 60.28 & 23.23 & 82.99 & 26.73 & 34.53 & 0.00 & 43.55 & 29.92 & 45.61\vartn{0.84} \\
    \cmidrule{2-21}
    \texttt{M} & 86.65 & 41.65 & 78.67 & 26.91 & 30.88 & 45.91 & 46.50 & 61.48 & 81.84 & 38.79 & 94.09 & 68.65 & 33.59 & 84.52 & 40.90 & 42.40 & 10.15 & 41.20 & 35.24 & 52.11\vartn{0.17}\\
    \cmidrule{2-21}
    \texttt{S} & 60.55 & 49.61 & 82.63 & 7.80 & 5.42 & 29.23 & 15.71 & 15.26 & 68.18 & 0.00 & 90.83 & 61.59 & 12.09 & 61.29 & 0.00 & 35.23 & 0.00 & 32.75 & 22.19 & 34.23\vartn{0.53}\\
    \cmidrule{2-21}
    \texttt{AN} & 47.44 & 7.01 & 38.73 & 8.42 & 3.59 & 23.04 & 18.42 & 5.75 & 19.33 & 5.82 & 5.65 & 26.61 & 10.39 & 50.46 & 4.10 & 0.00 & 0.79 & 8.42 & 0.28 & 14.96\vartn{0.09}\\ 
    \cmidrule{2-21}
    \texttt{AS} & 66.93 & 10.15 & 62.17 & 33.95 & 22.10 & 35.26 & 51.20 & 35.57 & 73.12 & 20.72 & 77.55 & 52.02 & 0.63 & 71.62  & 21.20 & 1.14 & 12.36 &  47.32 & 9.71 & 37.09\vartn{0.83} \\
    \cmidrule{2-21}
    \texttt{AR} & 73.41 & 21.42 & 77.58 & 19.41 & 16.96 & 33.63 &  44.90 & 38.53 & 80.96 & 29.31 & 94.98 & 56.89 & 17.04 & 76.15 & 16.15 & 7.07 & 5.11 & 23.74 & 1.24 & 38.66\vartn{1.12} \\ 
    \cmidrule{2-21}
    \texttt{AF} & 77.61 & 31.99 & 76.42 & 28.84 & 10.30 & 31.42 & 52.92 & 29.99 & 68.09 & 24.55 & 92.03 & 54.52 & 34.91 & 68.21 & 26.07 & 11.02 & 1.44 & 7.71 & 36.78 & 40.25\vartn{0.71} \\
    \bottomrule

		\end{tabular}
  }
	\smallskip\caption{\textbf{ResNet-50 class-wise performance.} We report the performance of \method (IoU \%) trained on GTAV with ResNet-50 as backbone.
    }
	\label{tab:gta5_rn50}
\end{table*}

\begin{table*}[t!]
	\setlength{\tabcolsep}{0.007\linewidth}
        \renewcommand{\arraystretch}{1.2}
        \newcommand{\best}{\bf}
		\centering
		\resizebox{\linewidth}{!}{
	  \begin{tabular}{c c c c c c c c c c c c c c c c c c c c | c}
			\toprule
			 \makecell{Target\\eval.}
			& \rotatebox{90}{\textcolor{csroad}{\rule{0.5em}{1.5ex}} road}
			& \rotatebox{90}{\textcolor{csside}{\rule{0.5em}{1.5ex}} sidewalk}
			& \rotatebox{90}{\textcolor{csbuild}{\rule{0.5em}{1.5ex}} building} 
			& \rotatebox{90}{\textcolor{cswall}{\rule{0.5em}{1.5ex}} wall} 
			& \rotatebox{90}{\textcolor{csfence}{\rule{0.5em}{1.5ex}} fence} 
			& \rotatebox{90}{\textcolor{cspole}{\rule{0.5em}{1.5ex}} pole} 
			& \rotatebox{90}{\textcolor{cslight}{\rule{0.5em}{1.5ex}} traffic light} 
			& \rotatebox{90}{\textcolor{cssign}{\rule{0.5em}{1.5ex}} traffic sign} 
			& \rotatebox{90}{\textcolor{csveg}{\rule{0.5em}{1.5ex}} vegetation} 
			& \rotatebox{90}{\textcolor{csterrain}{\rule{0.5em}{1.5ex}} terrain} 
			& \rotatebox{90}{\textcolor{cssky}{\rule{0.5em}{1.5ex}} sky} 
			& \rotatebox{90}{\textcolor{csperson}{\rule{0.5em}{1.5ex}} person} 
			& \rotatebox{90}{\textcolor{csrider}{\rule{0.5em}{1.5ex}} rider} 
			& \rotatebox{90}{\textcolor{cscar}{\rule{0.5em}{1.5ex}} car} 
			& \rotatebox{90}{\textcolor{cstruck}{\rule{0.5em}{1.5ex}} truck} 
			& \rotatebox{90}{\textcolor{csbus}{\rule{0.5em}{1.5ex}} bus} 
			& \rotatebox{90}{\textcolor{cstrain}{\rule{0.5em}{1.5ex}} train} 
			& \rotatebox{90}{\textcolor{csmbike}{\rule{0.5em}{1.5ex}} motorcycle} 
			& \rotatebox{90}{\textcolor{csbike}{\rule{0.5em}{1.5ex}} bicycle} 
			& mIoU\%\\
			\midrule

    \texttt{C} & 90.25 & 44.78 & 84.54 & 31.71 & 31.48 & 44.17 & 45.45 & 35.78 & 87.17 & 35.58 & 84.30 & 69.62 & 20.48 & 86.87 & 31.11 & 44.38 & 7.73 & 34.06 & 30.46 & 49.47\vartn{0.36}\\
    \cmidrule{2-21}
    \texttt{B} & 86.79 & 41.47 & 79.53 & 16.67 & 41.27 & 39.41 & 42.31 & 33.35 & 78.64 & 36.86 & 91.03 & 60.32 & 23.73 & 81.51 & 31.13 & 25.99 & 0.00 & 45.78 & 25.73 & 46.40\vartn{0.50}\\
    \cmidrule{2-21}
    \texttt{M} & 78.71 & 38.89 & 81.85 & 26.56 & 40.22 & 47.32 & 49.27 & 62.19 & 82.68 & 41.54 & 95.63 & 67.60 & 25.87 & 85.50 & 41.62 & 35.87 & 12.55 & 43.69 & 29.92 & 51.97\vartn{1.30} \\
    \cmidrule{2-21}
    \texttt{S} & 63.72 & 56.80 & 85.05 & 9.30 & 21.62 & 33.26 & 16.44 & 18.96 & 69.42 & 0.00 & 92.10 & 63.52 & 10.95 & 64.86 & 0.00 & 29.84 & 0.00 & 39.67 & 22.22 & 36.72\vartn{0.71}\\
    \cmidrule{2-21}
    \texttt{AN} & 65.87 & 23.23 & 37.83 & 13.72 & 4.60 & 30.34 & 16.49 & 7.48 & 27.37 & 7.83 & 17.61 & 35.16 & 18.48 & 53.71 & 5.67 & 0.00 & 0.84 & 10.25 & 1.44 & 19.89\vartn{1.22}\\ 
    \cmidrule{2-21}
    \texttt{AS} & 75.42 & 31.90 & 72.15 & 36.75 & 27.85 & 38.89 & 49.63 & 33.09 & 72.45 & 22.98 & 84.21 & 56.75 & 1.91 & 75.84 & 34.61 & 4.44 & 4.17 & 48.91 & 14.26 & 41.38\vartn{0.34}  \\
    \cmidrule{2-21}
    \texttt{AR} & 57.58 & 26.76 & 79.76 & 19.79 & 21.06 & 37.70 & 46.34 & 37.66 & 83.85 & 36.96 & 94.80 & 55.40 & 31.61 & 79.53 & 14.84 & 14.01 & 6.97 & 29.36 & 3.34 & 40.91\vartn{1.28} \\ 
    \cmidrule{2-21}
    \texttt{AF} & 77.96 & 41.73 & 77.99 & 34.73 & 6.85 & 36.80 & 49.49 & 34.51 & 72.00 & 32.60 & 91.52 & 46.28 & 27.28 & 70.77 & 31.17 & 19.08 & 4.87 & 10.75 & 34.55 & 42.15\vartn{1.87} \\
    \bottomrule

		\end{tabular}
  }
	\smallskip\caption{\textbf{ResNet-101 class-wise performance.} We report the performance of \method (IoU \%) trained on GTAV with ResNet-101 as backbone.
    }
	\label{tab:gta5_rn101}
\end{table*}

\section{Prompts used for style mining}
\label{sec:prompts_used}
The $<$\texttt{random style prompt}$>$ used for training \method:\\
$\mathcal{R}_1$ = $<$\texttt{random style prompt}$>$ = \{ \textit{Ethereal Mist, Cyberpunk Cityscape, Rustic Charm, Galactic Fantasy, Pastel Dreams, Dystopian Noir, Whimsical Wonderland, Urban Grit,
Enchanted Forest, Retro Futurism, Monochrome Elegance, Vibrant Graffiti, Haunting Shadows, Steampunk Adventures, Watercolor Serenity, Industrial Chic, Cosmic Voyage, Pop Art Popularity, Abstract Symphony, Magical Realism, Abstract Geometric Patterns,
    Vintage Film Grain,
    Neon Cityscape Vibes,
    Surreal Watercolor Dreams,
    Minimalist Nature Scenes,
    Cyberpunk Urban Chaos,
    Impressionist Sunset Hues,
    Pop Art Explosion,
    Fantasy Forest Adventures,
    Pixelated Digital Chaos,
    Monochromatic Street Photography,
    Vibrant Graffiti Expressions,
    Steampunk Industrial Charm,
    Ethereal Cloudscapes,
    Retro Futurism Flare,
    Dark and Moody Landscapes,
    Pastel Dreamworlds,
    Galactic Space Odyssey,
    Abstract Brush Strokes,
    Noir Cinematic Moments,
    Whimsical Fairy Tale Realms,
    Modernist Architectural Wonders,
    Macro Botanical Elegance,
    Dystopian Sci-Fi Realities,
    High Contrast Street Art,
    Impressionist City Reflections,
    Pixel Art Nostalgia,
    Dynamic Action Sequences,
    Soft Focus Pastels,
    Abstract 3D Renderings,
    Mystical Moonlit Landscapes,
    Urban Decay Aesthetics,
    Holographic Futuristic Visions,
    Vintage Polaroid Snapshots,
    Digital Glitch Anomalies,
    Japanese Zen Gardens,
    Psychedelic Kaleidoscopes,
    Cosmic Abstract Portraits,
    Subtle Earthy Textures,
    Hyperrealistic Wildlife Portraits,
    Cybernetic Neon Lights,
    Warped Reality Illusions,
    Whimsical Watercolor Animals,
    Industrial Grunge Textures,
    Tropical Paradise Escapes,
    Dynamic Street Performances,
    Abstract Architectural Wonders,
    Comic Book Panel Vibes,
    Soft Glow Sunsets,
    8-Bit Pixel Adventures,
    Galactic Nebula Explosions,
    Doodle Sketchbook Pages,
    High-Tech Futuristic Landscapes,
    Cinematic Noir Shadows,
    Vibrant Desert Landscapes,
    Abstract Collage Chaos,
    Nature in Infrared,
    Surreal Dream Sequences,
    Abstract Light Painting,
    Whimsical Fantasy Creatures,
    Cybernetic Augmented Reality,
    Impressionist Rainy Days,
    Vintage Aged Photographs,
    Neon Anime Cityscapes,
    Pastel Sunset Palette,
    Surreal Floating Islands,
    Abstract Mosaic Patterns,
    Retro Sci-Fi Spaceships,
    Futuristic Cyber Landscapes,
    Steampunk Clockwork Contraptions,
    Monochromatic Urban Decay,
    Glitch Art Distortions,
    Magical Forest Enchantments,
    Digital Oil Painting,
    Pop Surrealist Dreams,
    Dynamic Graffiti Murals,
    Vintage Pin-up Glamour,
    Abstract Kinetic Sculptures,
    Neon Jungle Adventures,
    Minimalist Futuristic Interfaces} \} \\

\noindent The $<$\texttt{random character prompt}$>$ used in~\cref{tab:class_name_ablation} experiments (\ie, `RCP') are:\\
$\mathcal{R}_2$ = $<$\texttt{random character prompt}$>$ = \{ \textit{ioscjspa,
cjosae,
wqvsecpas,
csavwggw,
csanoiaj,
zfaspf,
atpwqkmfc,
mdmfejh,
casjicjai,
cnoacpoaj,
noiasvnai,
kcsakofnaoi,
cjncioasn,
wkqgmdc,
jqblhyu,
pqwfkgr,
mzxanqnw,
wnzsalml,
sdqlhkjr,
odfeqfit}
\} \\

\noindent Both $\mathcal{R}_1$ and $\mathcal{R}_2$ are concatenated with the word \texttt{style}.

\section{Limitations and perspectives}
\label{sec:limit}

\subsection{Limitations}
\label{sec:fail}

\begin{figure*}[h]
	
	\scriptsize	
	\setlength{\tabcolsep}{0.002\linewidth}
	\centering
        \tiny
	\begin{tabular}{cccc}
				{\scriptsize Image} & {\scriptsize GT} & {\scriptsize Baseline} & {\scriptsize\method}\\
		  \includegraphics[width=0.248\linewidth]{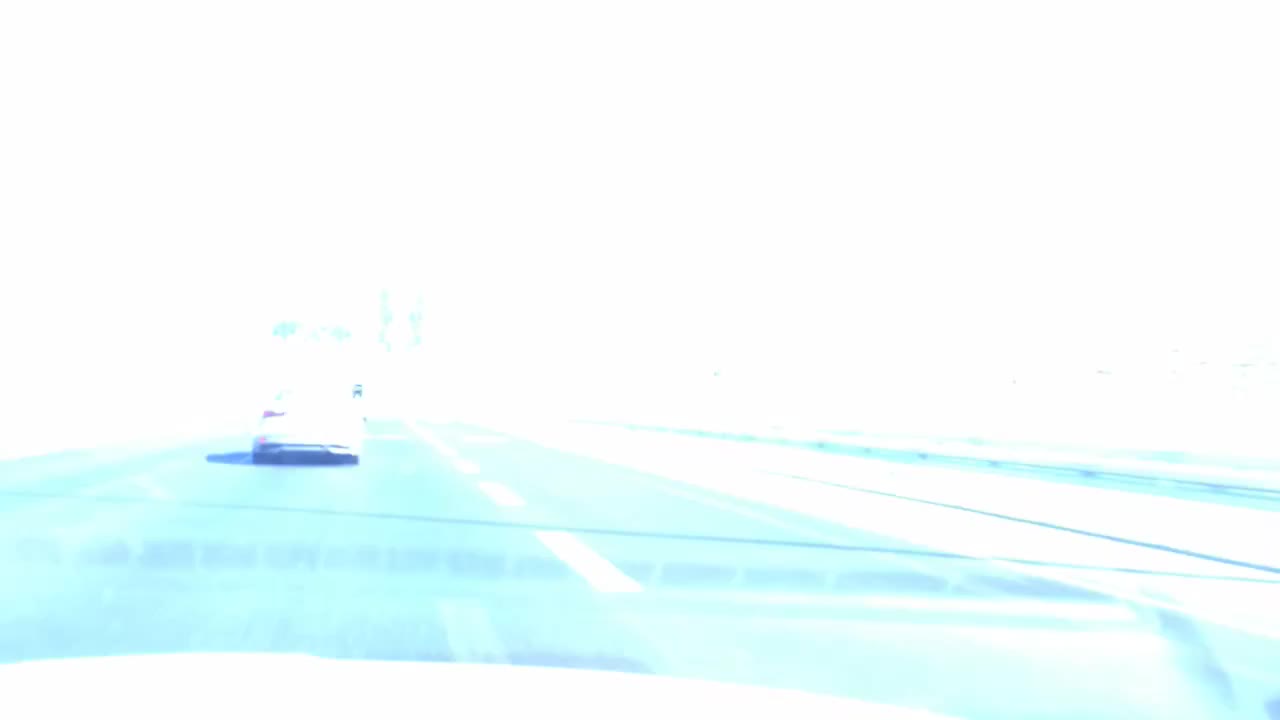}   
       &  \includegraphics[width=0.248\linewidth]{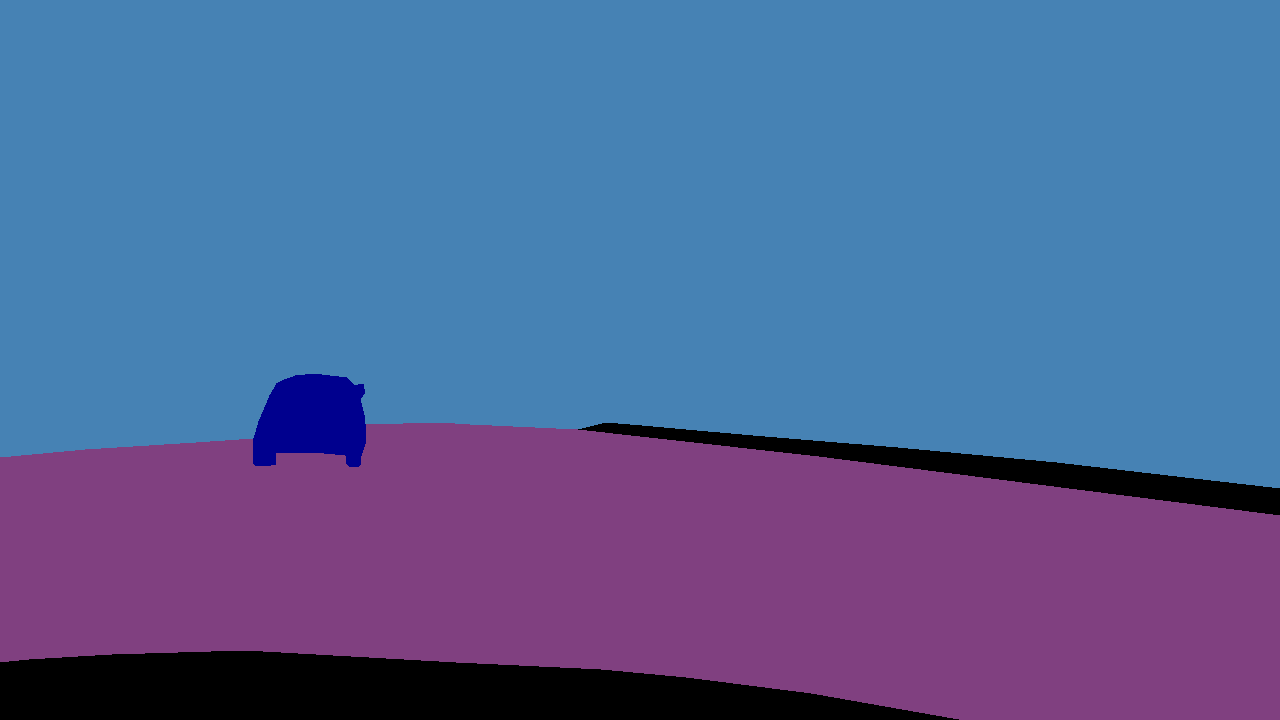}  
       & \includegraphics[width=0.248\linewidth]{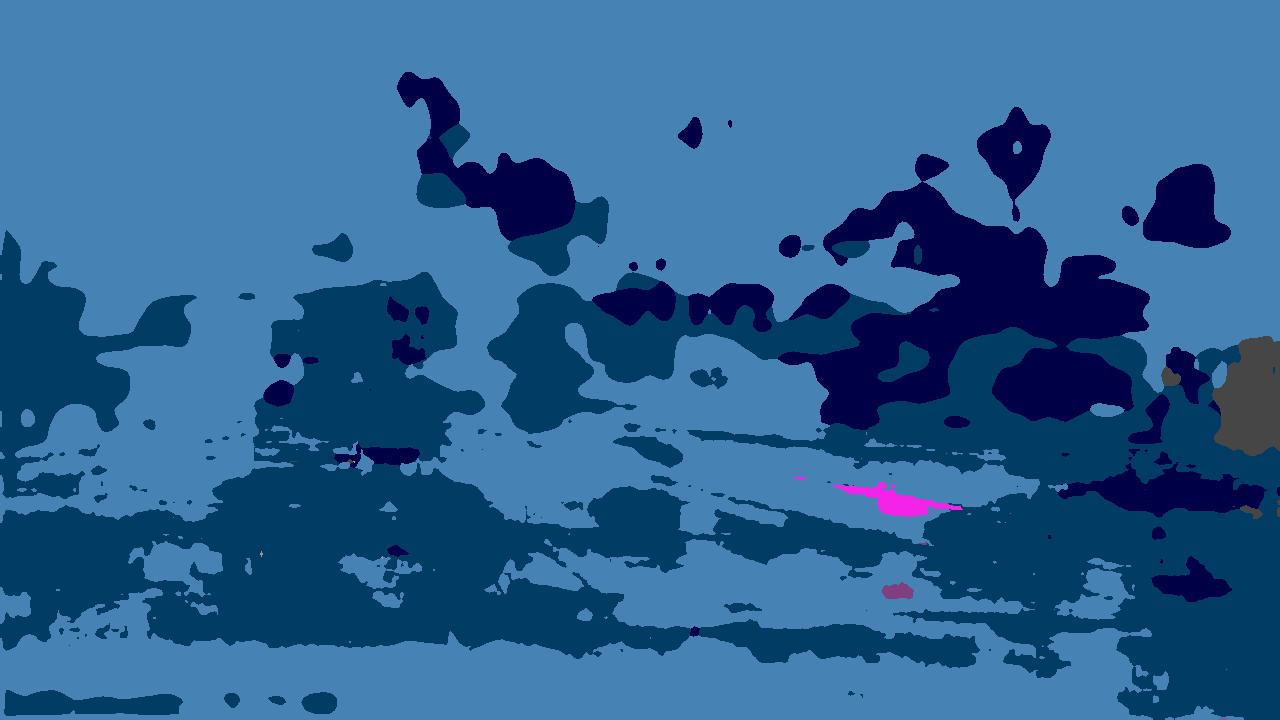}
      & \includegraphics[width=0.248\linewidth]{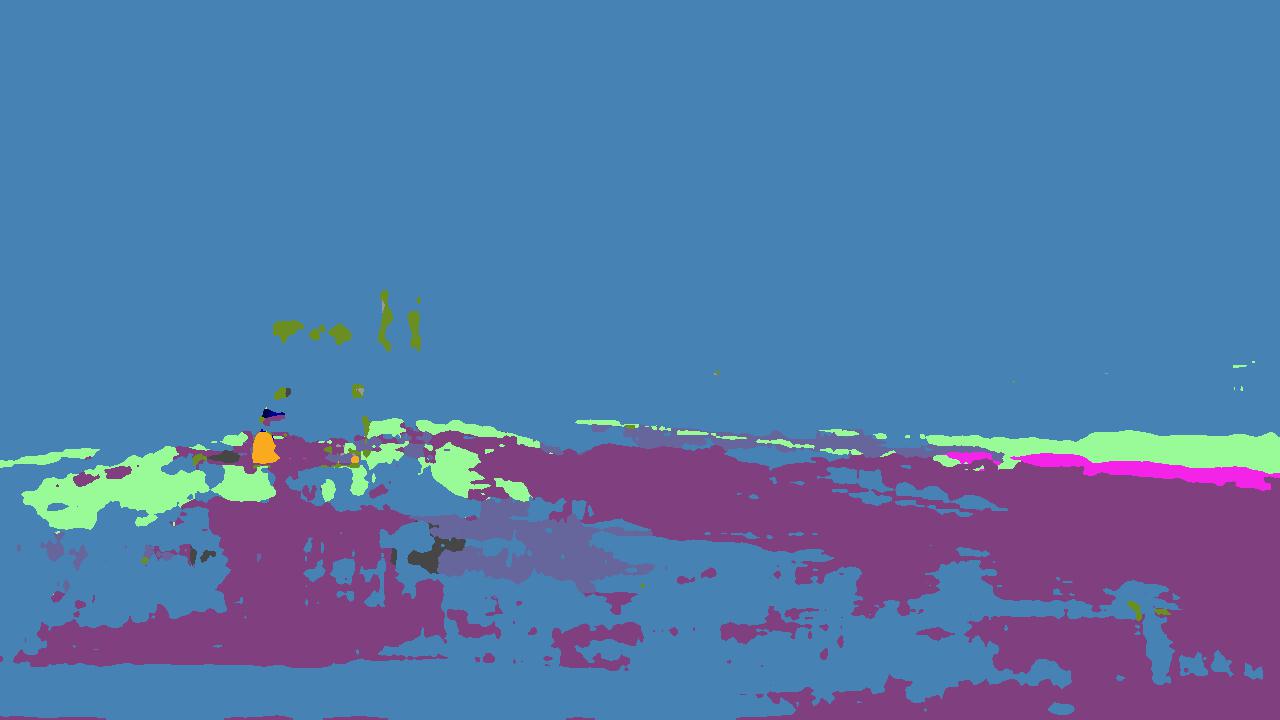} \\

      \includegraphics[width=0.248\linewidth]{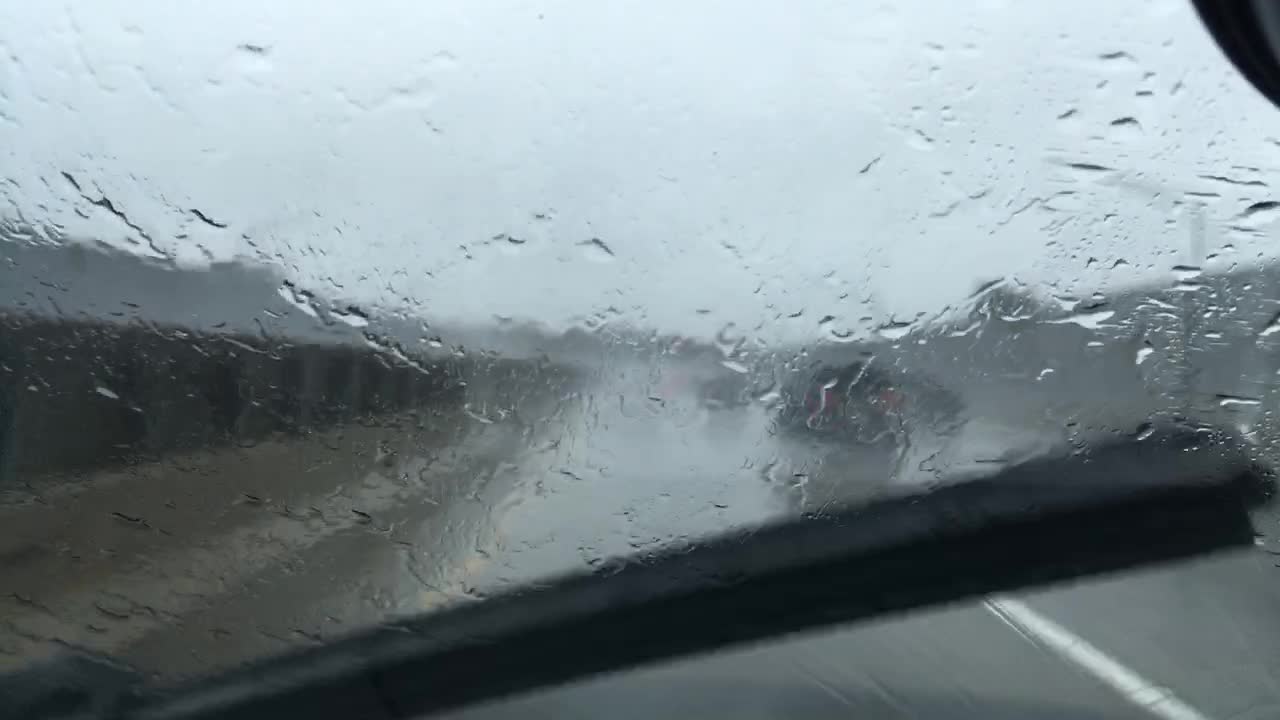}
      
       &  \includegraphics[width=0.248\linewidth]{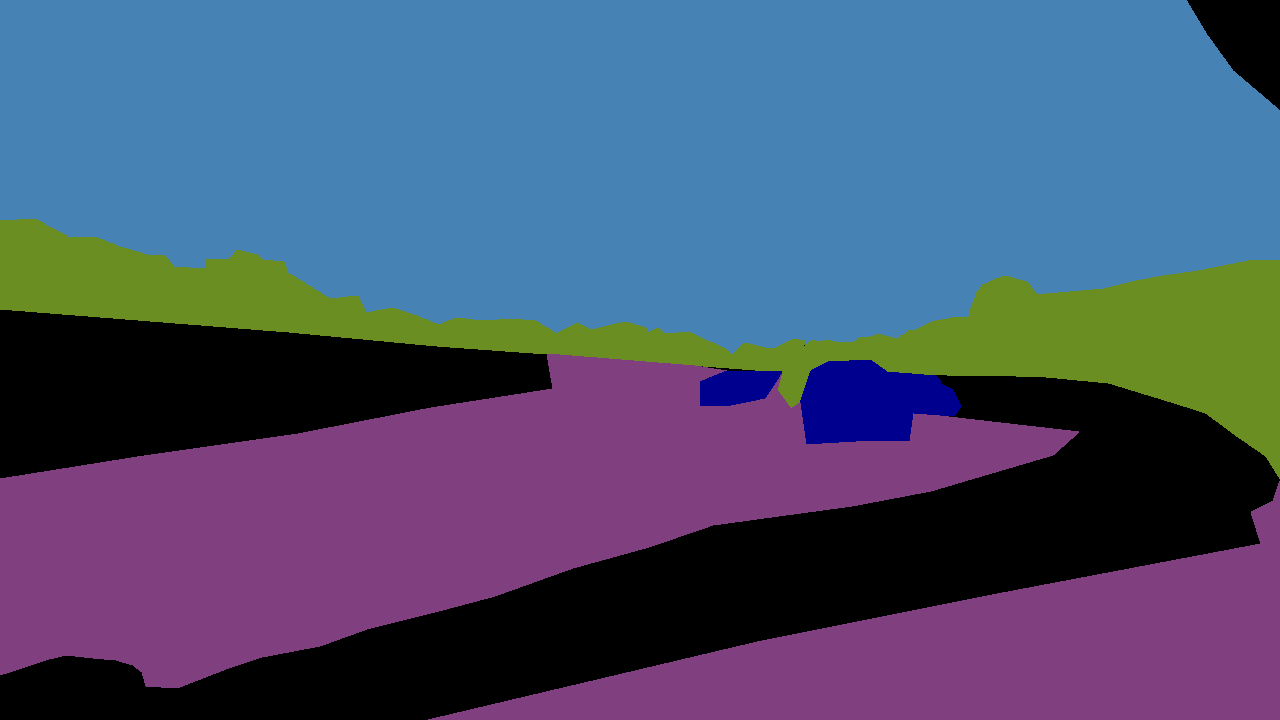}  
       & \includegraphics[width=0.248\linewidth]{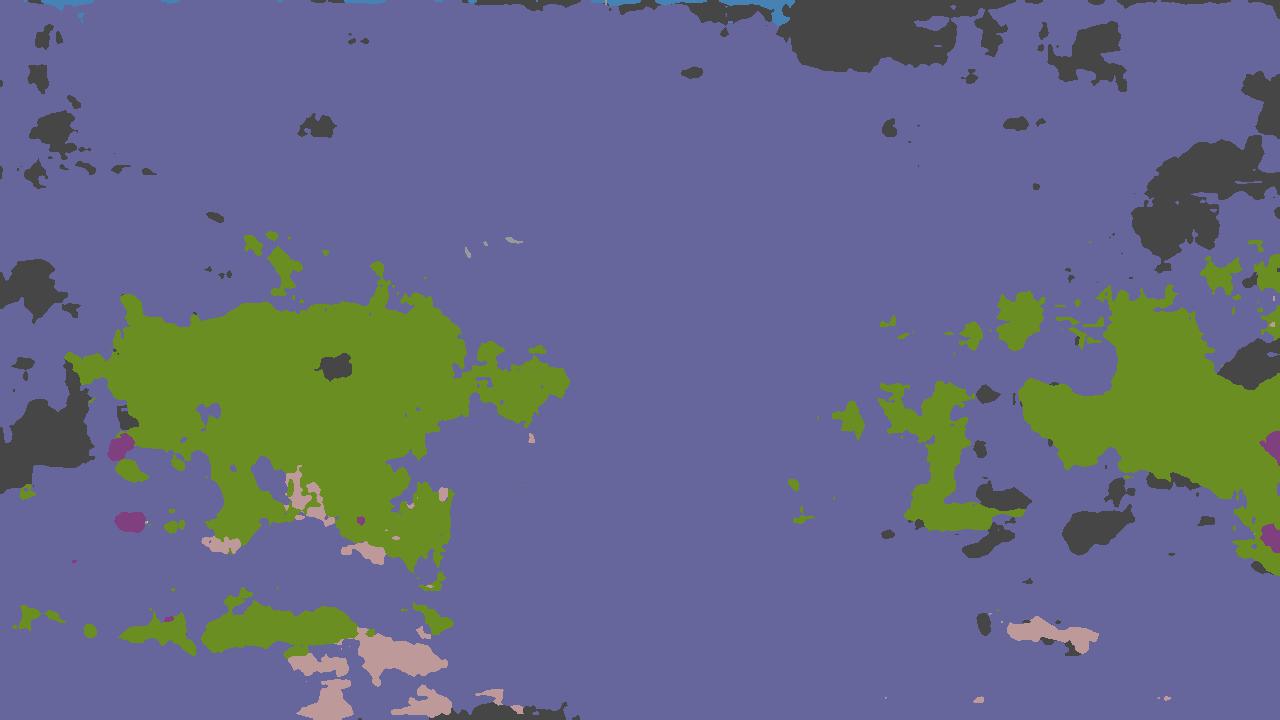}
      & \includegraphics[width=0.248\linewidth]{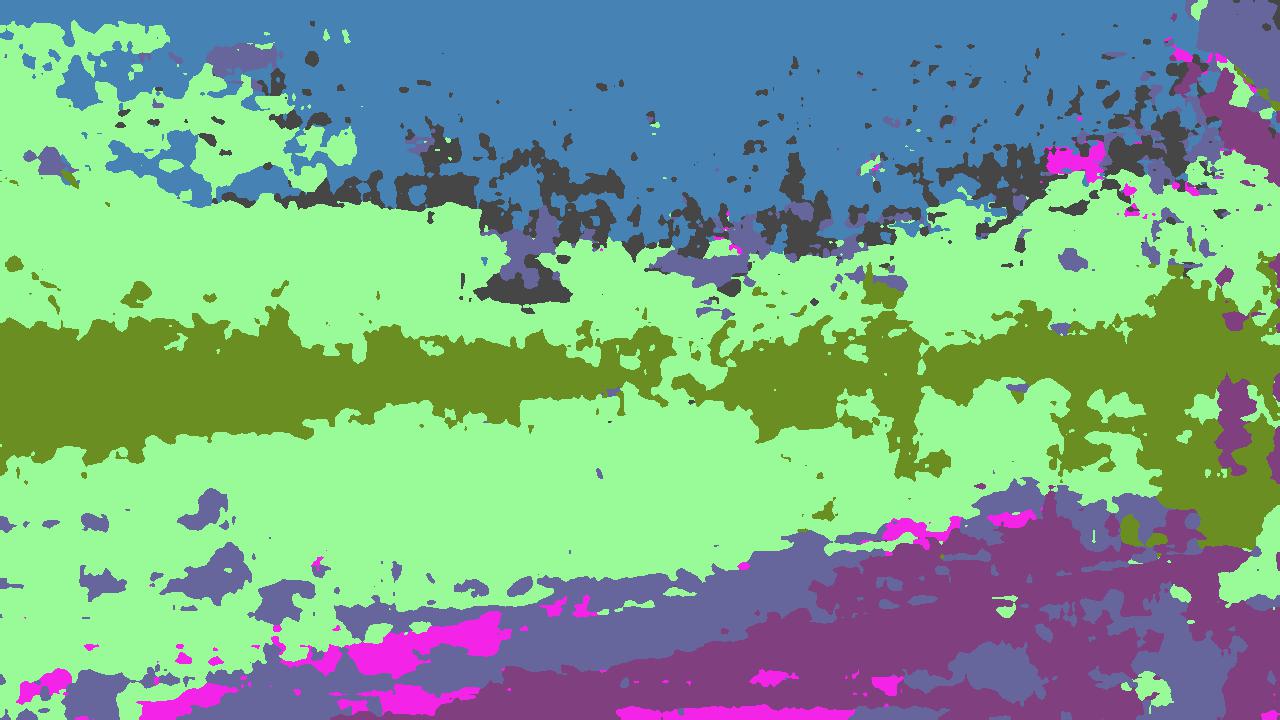} \\

      \includegraphics[width=0.248\linewidth]{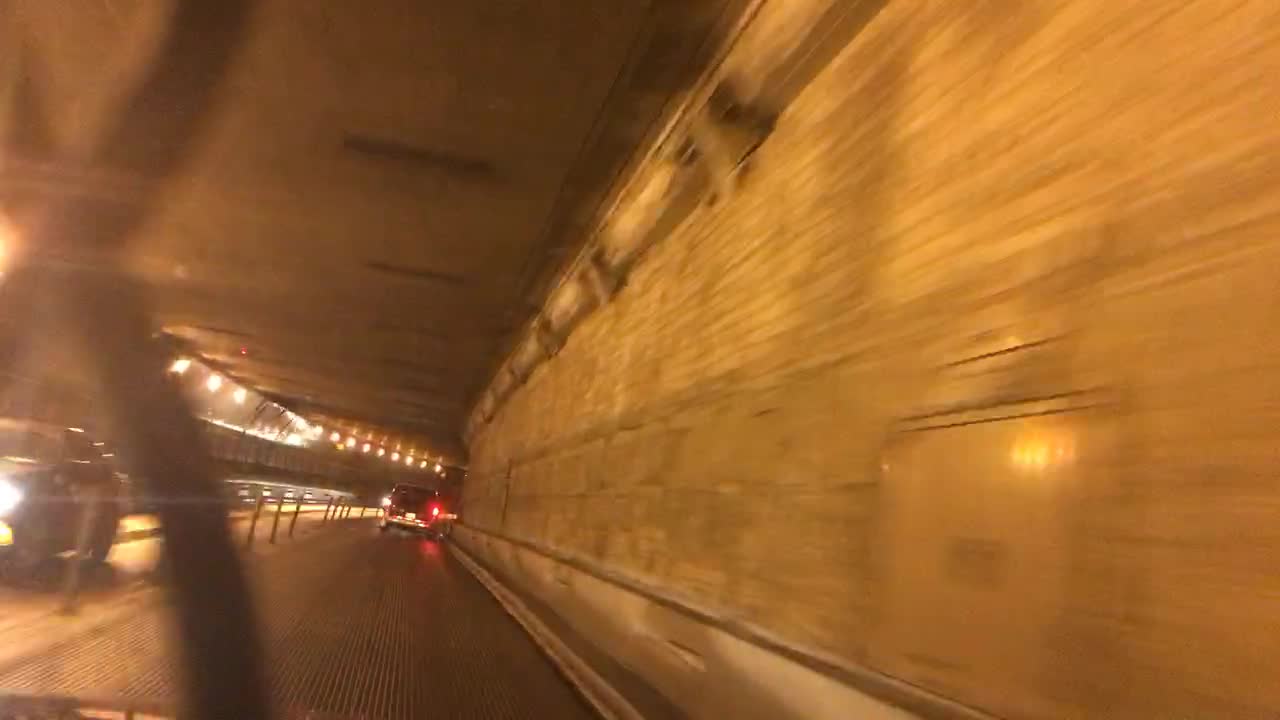}
      
       &  \includegraphics[width=0.248\linewidth]{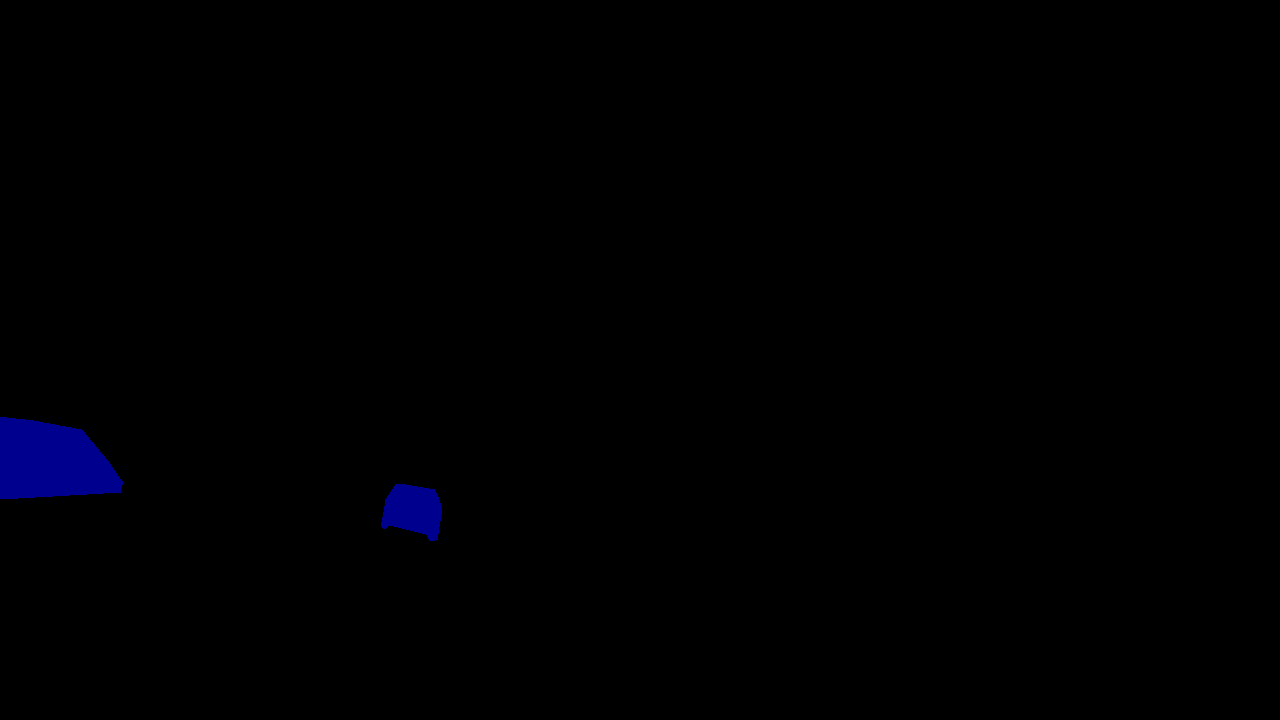}  
       & \includegraphics[width=0.248\linewidth]{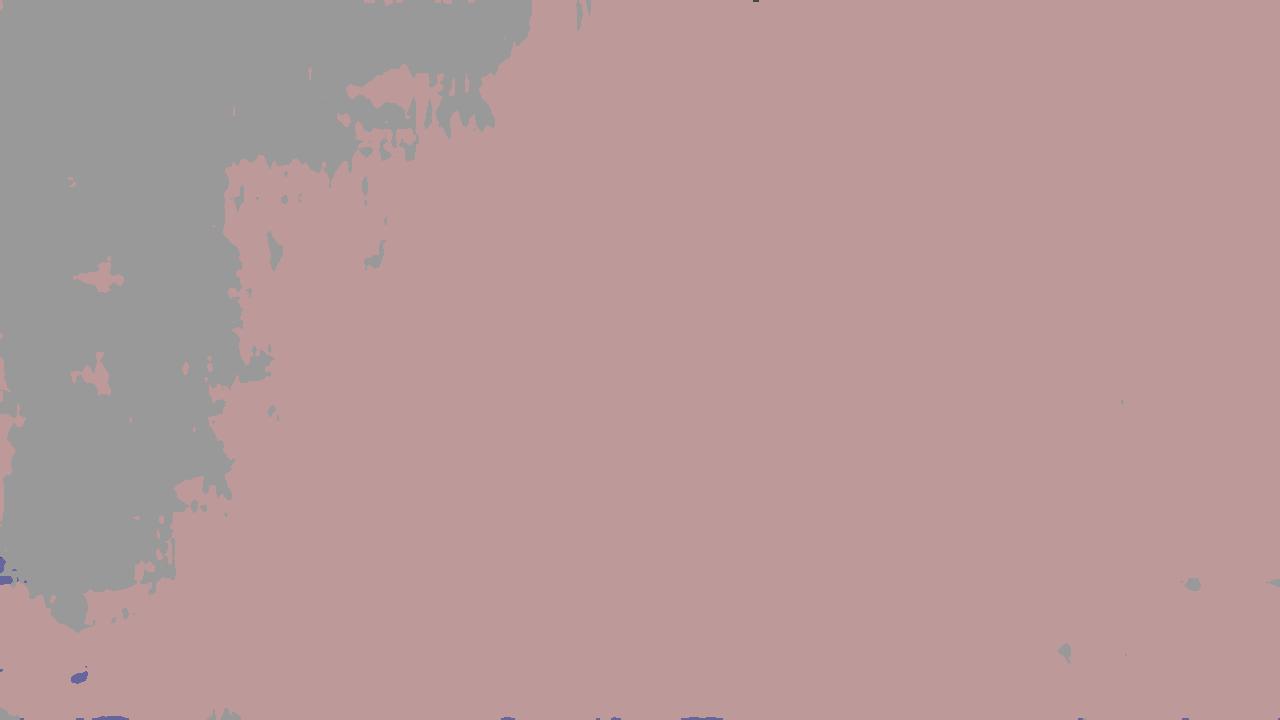}
      & \includegraphics[width=0.248\linewidth]{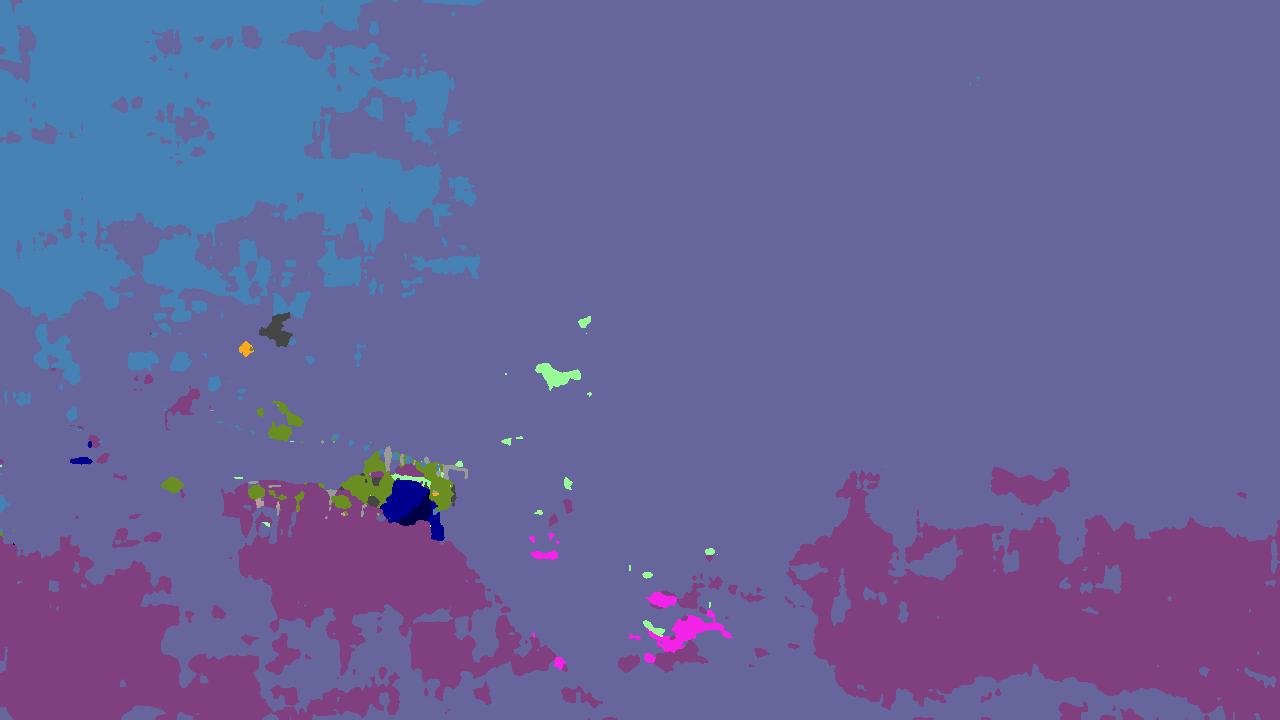} \\
    
    \includegraphics[width=0.248\linewidth]{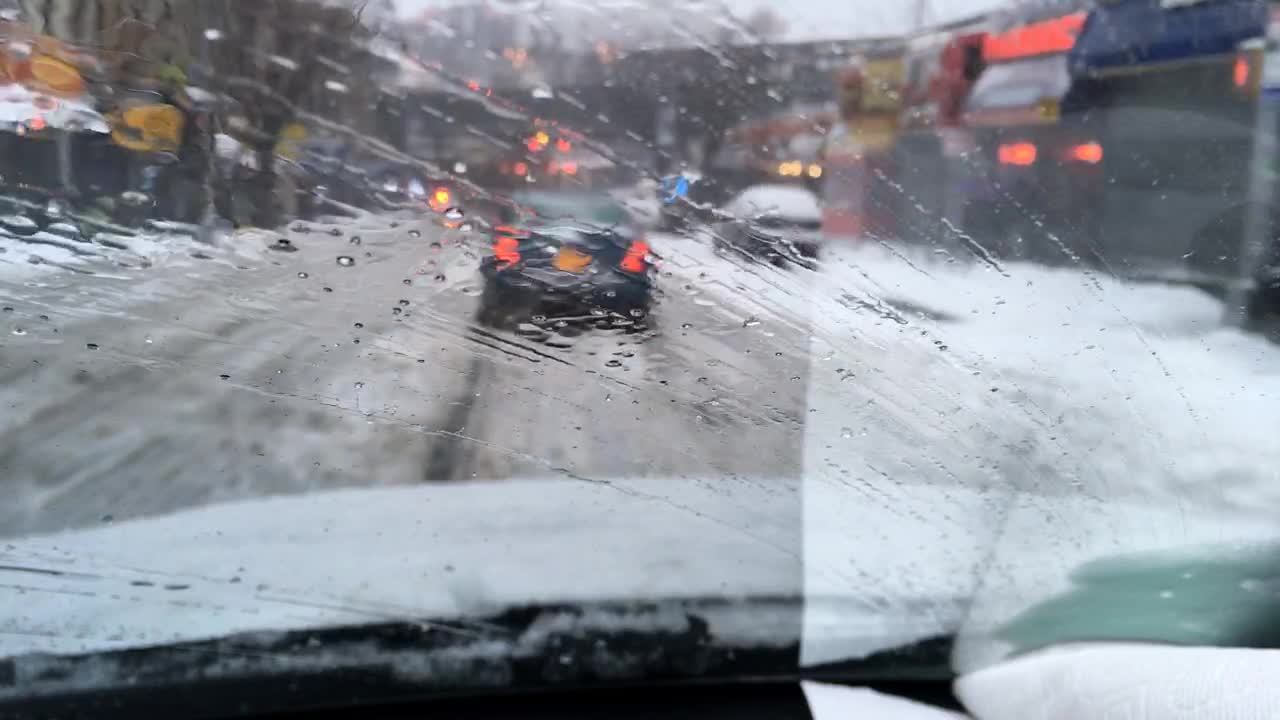}   
   
       &  \includegraphics[width=0.248\linewidth]{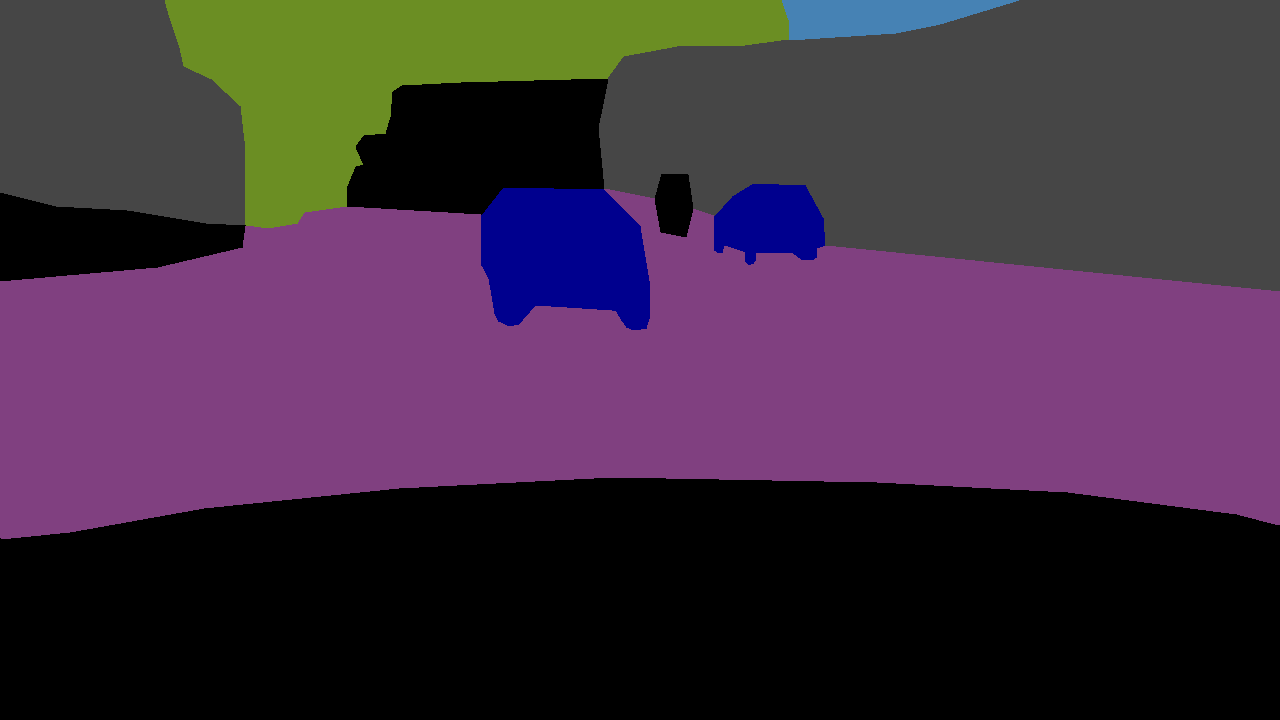}  
         & \includegraphics[width=0.248\linewidth]{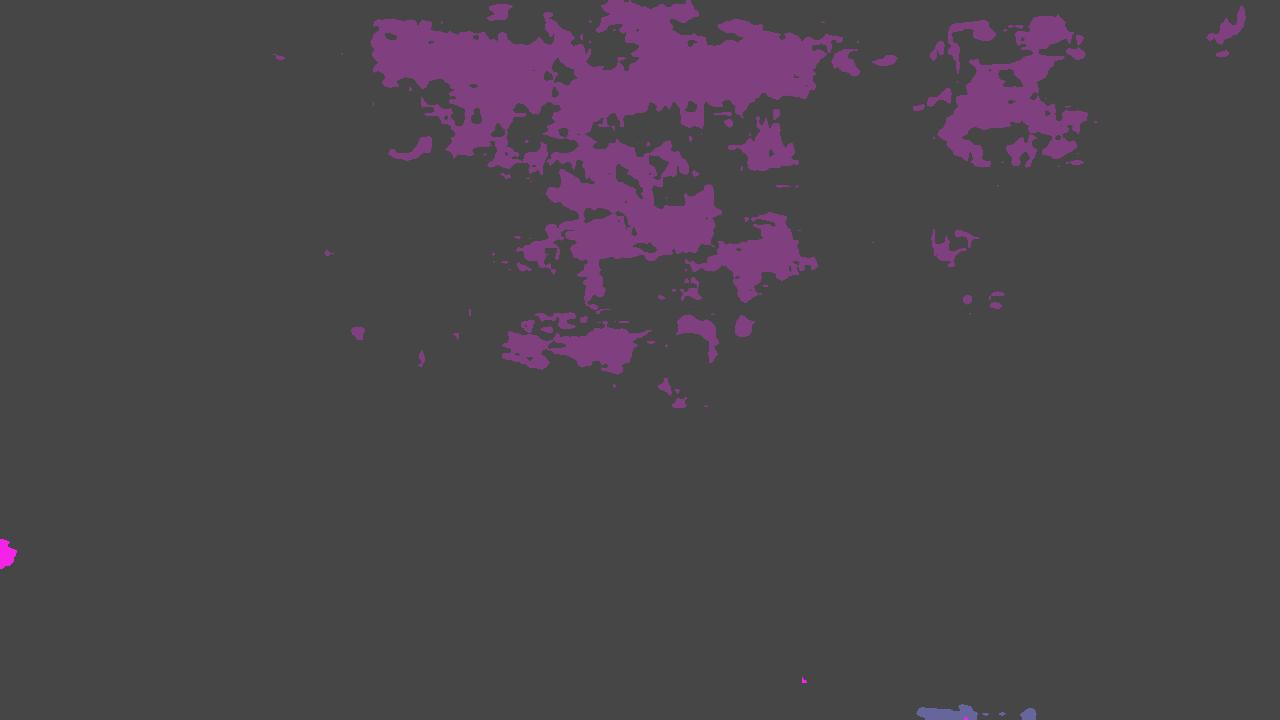}
      & \includegraphics[width=0.248\linewidth]{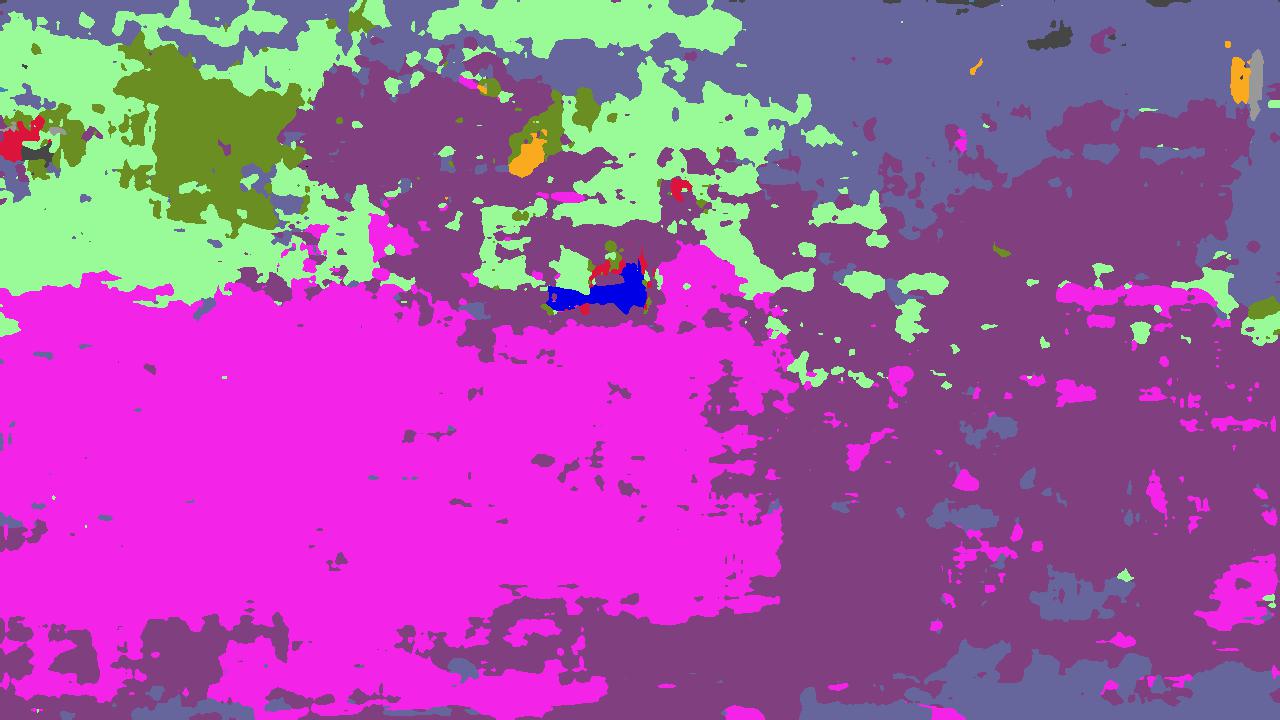} \\

        \includegraphics[width=0.248\linewidth]{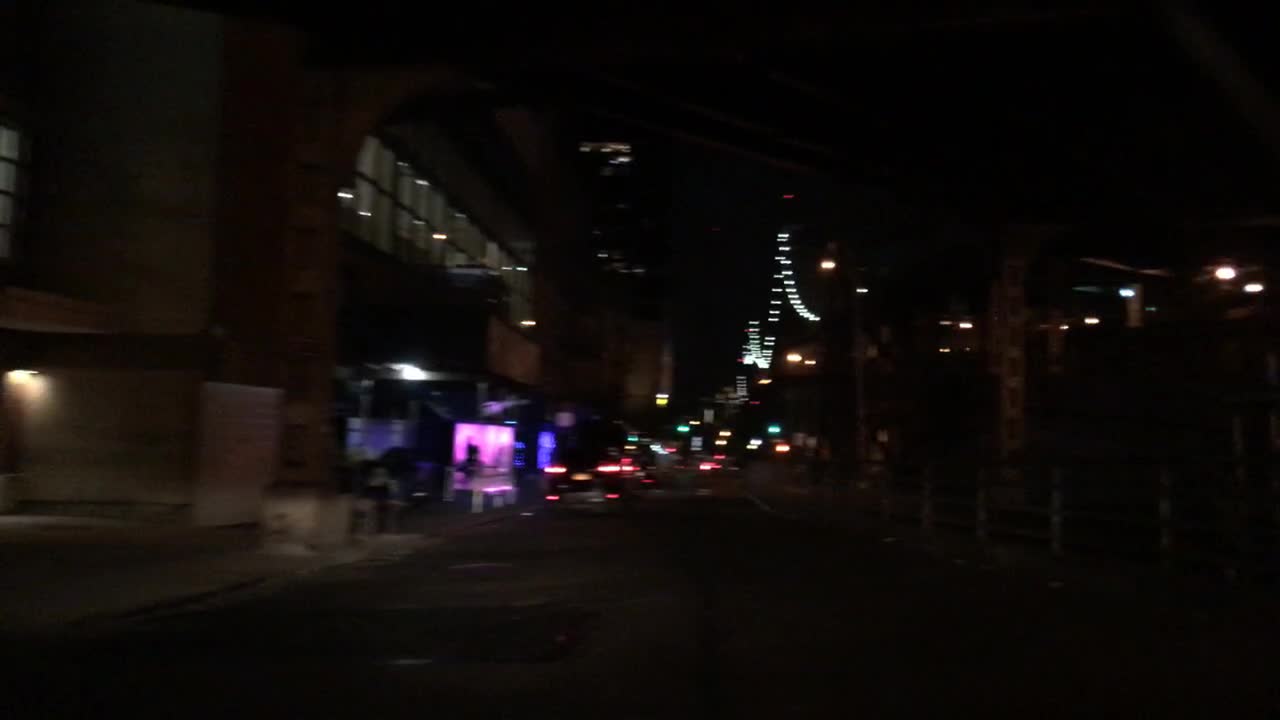}   
       
       &  \includegraphics[width=0.248\linewidth]{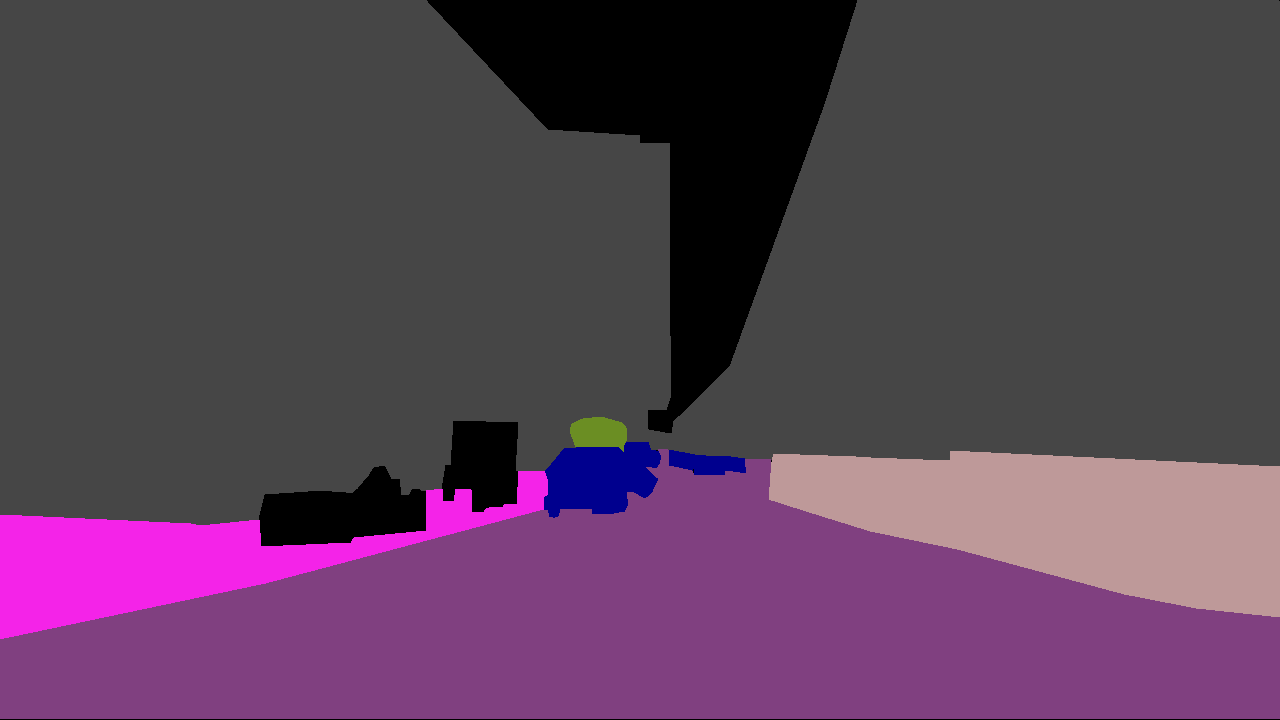}  
         & \includegraphics[width=0.248\linewidth]{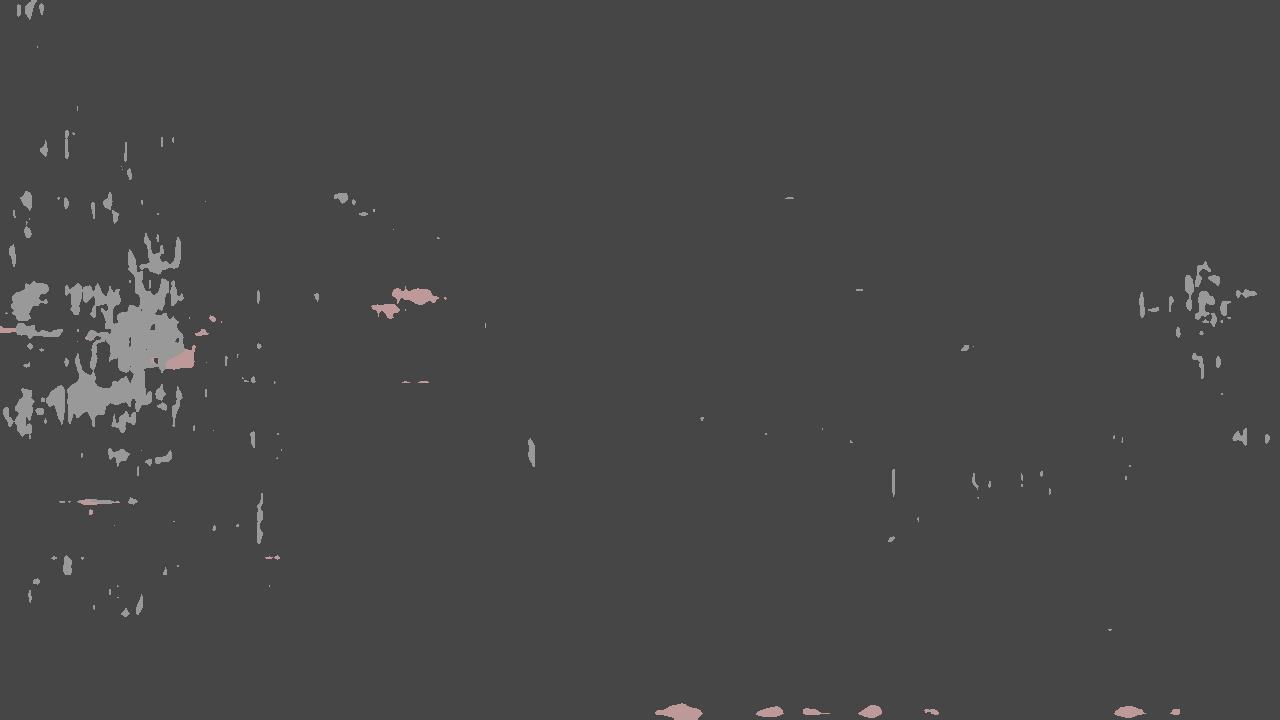}
      & \includegraphics[width=0.248\linewidth]{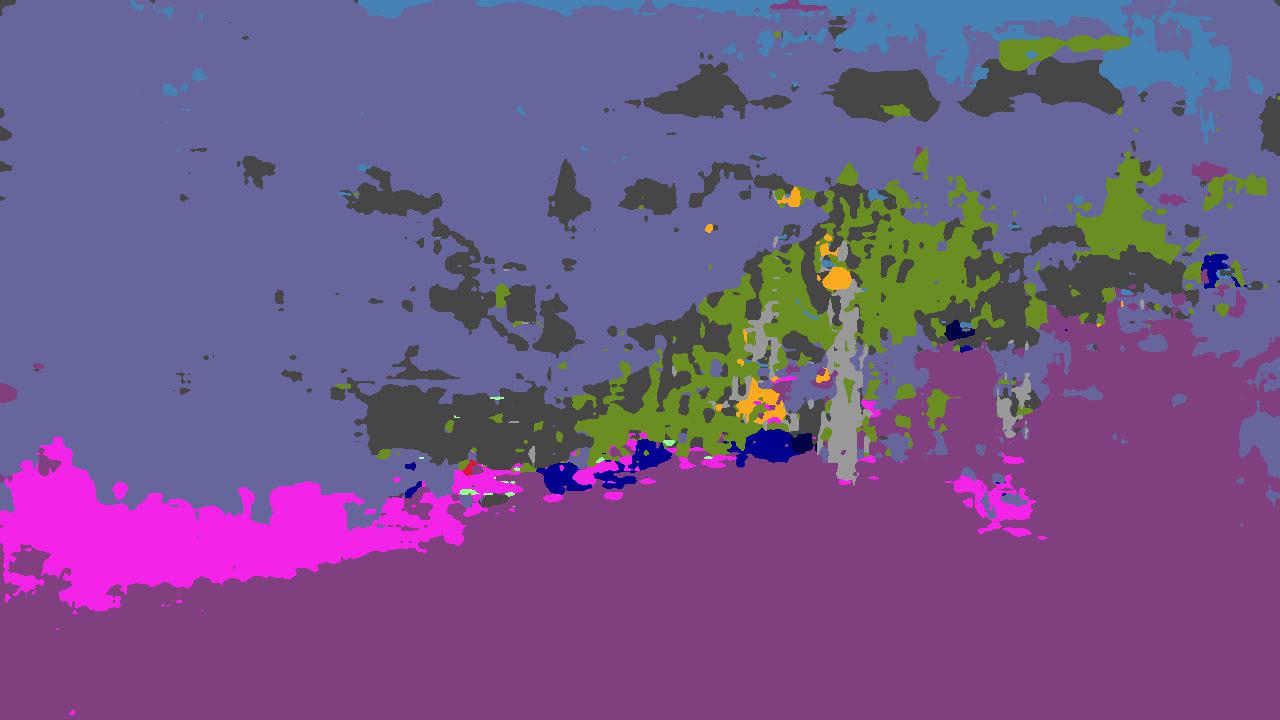} \\

        \includegraphics[width=0.248\linewidth]{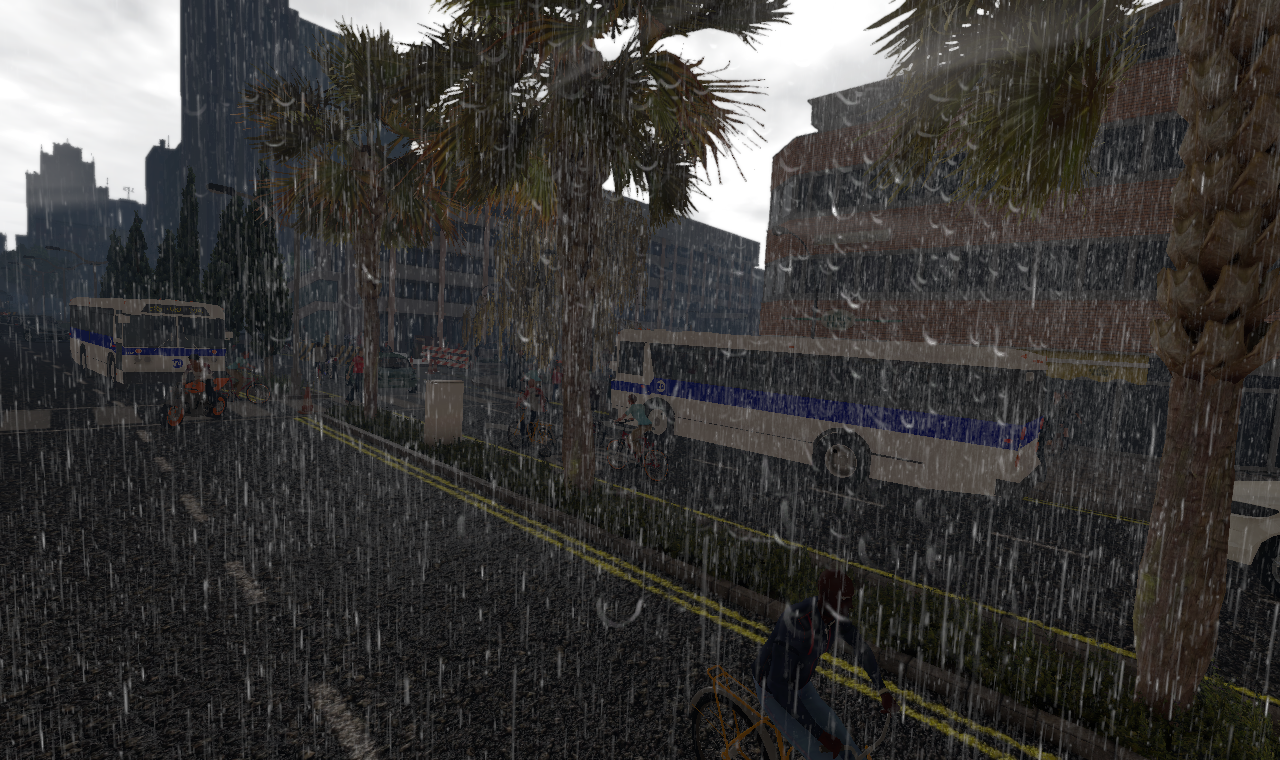}   
      
       &  \includegraphics[width=0.248\linewidth]{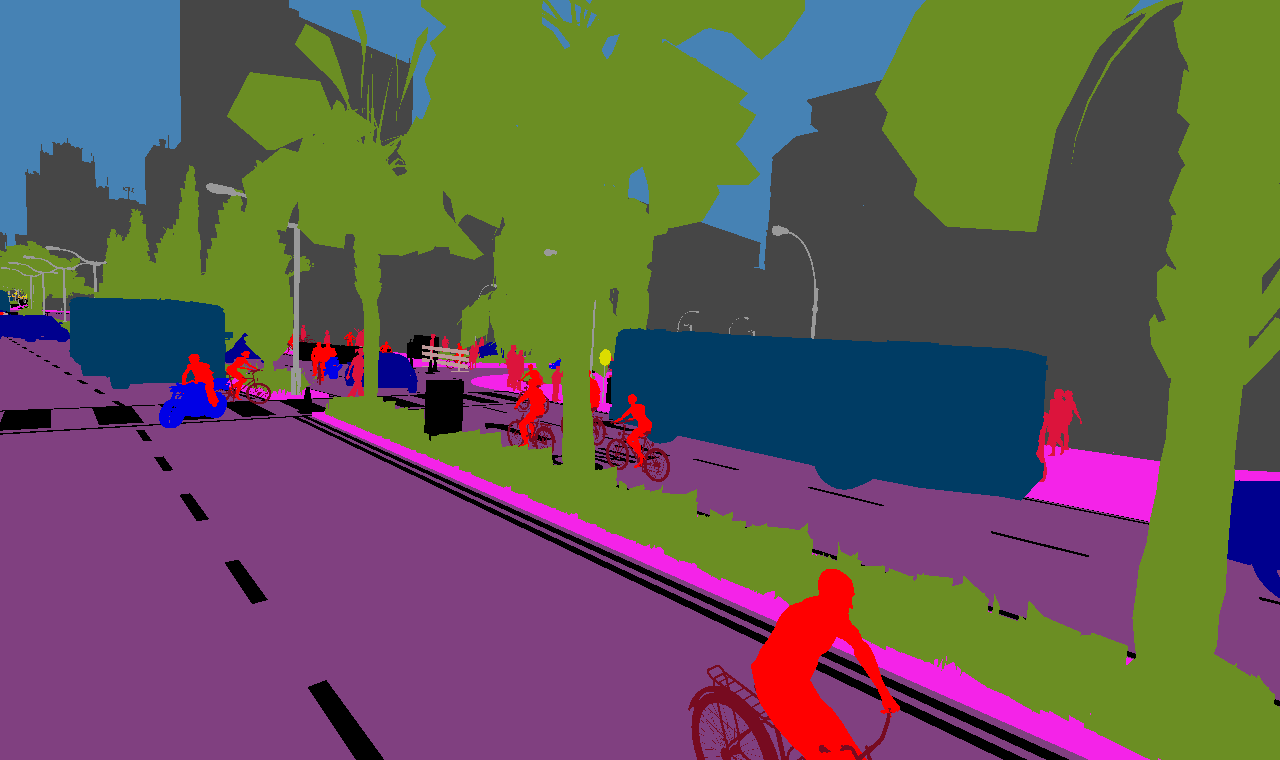}  
          & \includegraphics[width=0.248\linewidth]{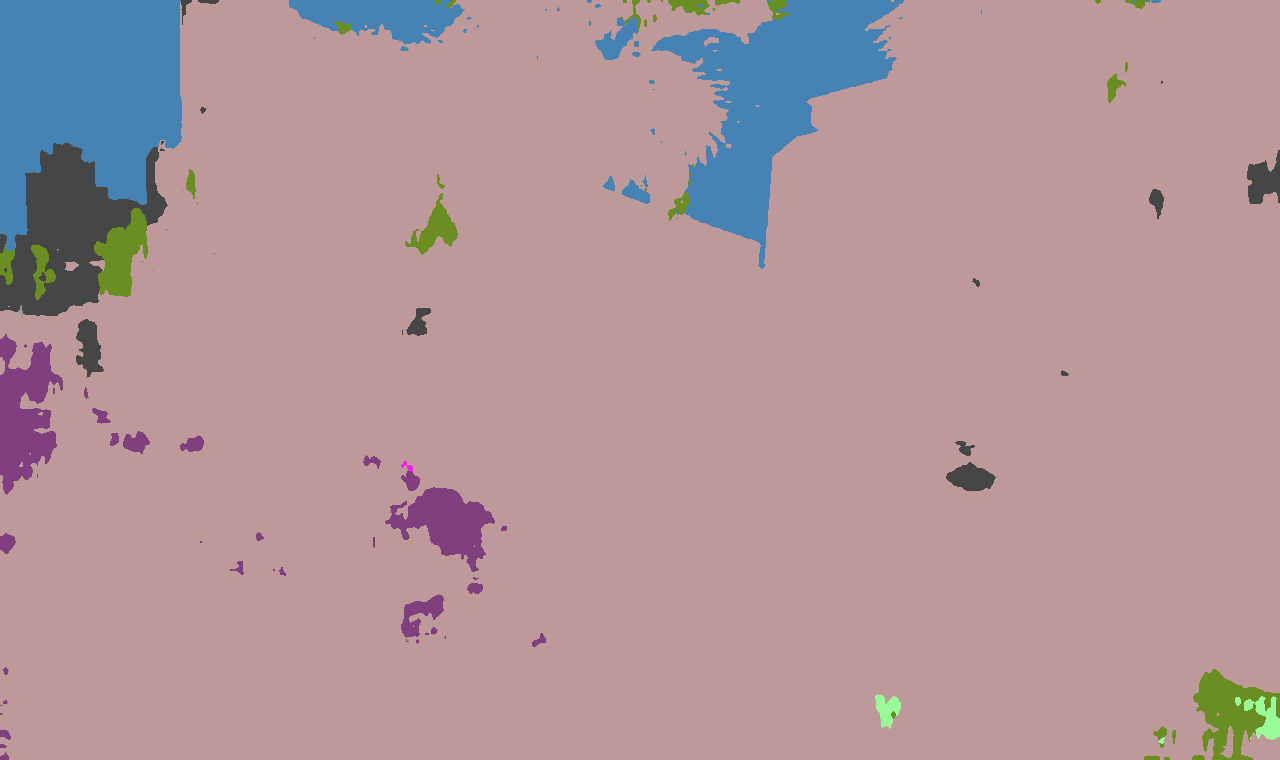}
      & \includegraphics[width=0.248\linewidth]{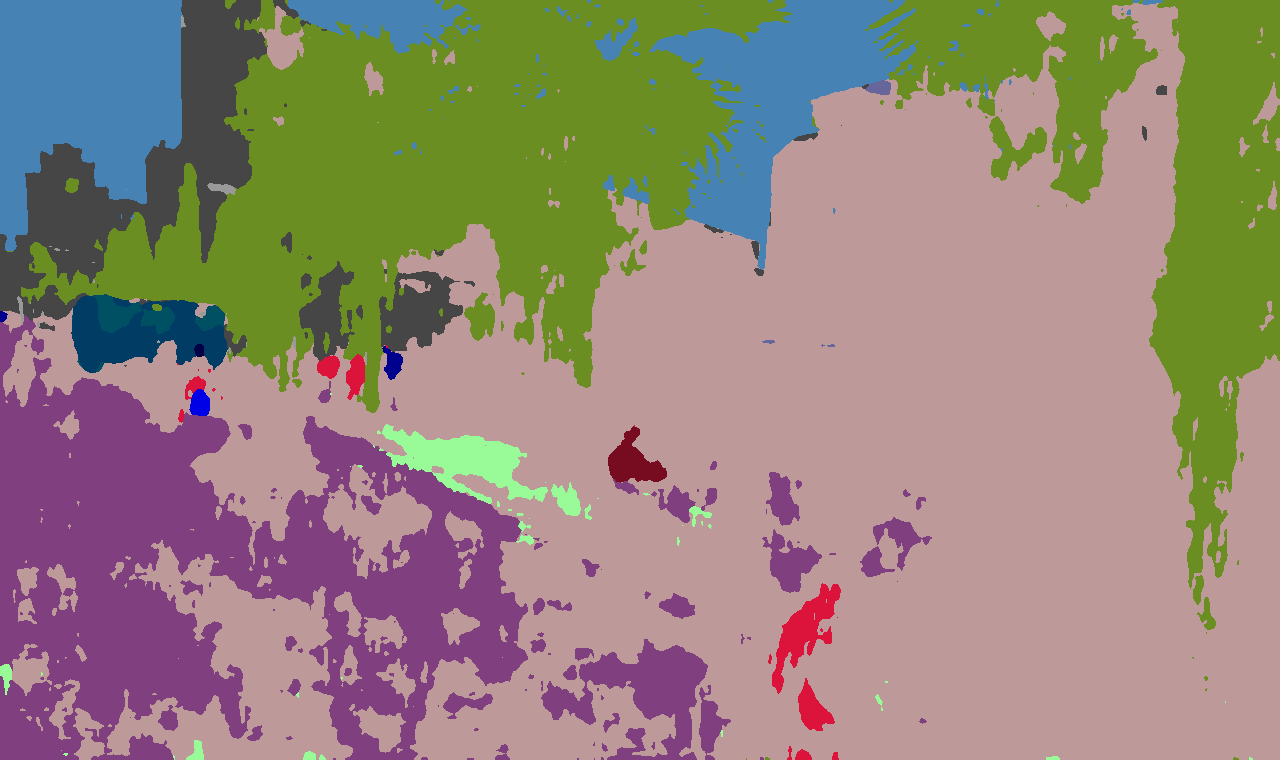} \\

	\end{tabular}
    \newcolumntype{C}[1]{>{\centering\let\newline\\\arraybackslash\hspace{0pt}}m{#1}}
    \begin{tabular}{C{5em}C{5em}C{5em}C{5em}C{5em}C{5em}C{5em}C{5em}C{5em}C{5em}}
    \cellcolor{csroad}{\textcolor{white}{Road}} & \cellcolor{csside}{\textcolor{black}{Sidewalk}} & \cellcolor{csbuild}{\textcolor{white}{Building}} & \cellcolor{cswall}{\textcolor{white}{Wall}} & \cellcolor{csfence}{\textcolor{black}{Fence}} & \cellcolor{cspole}{\textcolor{black}{Pole}} & \cellcolor{cslight}{\textcolor{black}{Traffic light}} & \cellcolor{cssign}{\textcolor{black}{Traffic sign}} & \cellcolor{csveg}{\textcolor{black}{Vegetation}} & \cellcolor{csterrain}{\textcolor{black}{Terrain}} \\
    \cellcolor{cssky}{\textcolor{white}{Sky}} & \cellcolor{csperson}{\textcolor{white}{Person}} & \cellcolor{csrider}{\textcolor{white}{Rider}} & \cellcolor{cscar}{\textcolor{white}{Car}} & \cellcolor{cstruck}{\textcolor{white}{Truck}} & \cellcolor{csbus}{\textcolor{white}{Bus}} & \cellcolor{cstrain}{\textcolor{white}{Train}} & \cellcolor{csmbike}{\textcolor{white}{Motorbike}} & \cellcolor{csbike}{\textcolor{white}{Bicycle}} & \cellcolor{csignore}{\textcolor{white}{n/a}} \\

    \end{tabular}
    \vspace{-0.1cm}
    \smallskip\caption{\textbf{Examples of failure cases.} \textit{Columns 1-2}: Image and Ground Truth (GT), \textit{Column 3}: Baseline (Freeze \xmark, Augment \xmark, Mix \xmark), \textit{Column 4}: \method results. The models are trained on GTAV with ResNet-50 backbone.}
    \vspace{-0.3cm}
    \label{fig:qual_failure}
\end{figure*}

\paragraph{Failures conditions.} We show in~\cref{fig:qual_failure} failure cases in rare conditions, which include extreme illumination or darkness, low visibility due to rain drops on the windshield, and other adverse conditions (\eg, snowy road). While \method improves over the baseline, the results remain unsatisfactory for safety-critical applications as it fails to segment critical objects in the scenes (\eg, car, road, sidewalk, person etc). 
We leave for future research the generalization to the above mentioned conditioned. One possible direction could be to design specific methods for specific corner conditions (\eg,~\cite{halder2019physics,lengyel2021zero}), although we highlight this is orthogonal to generalization.

\paragraph{Stylization.}At the heart of \method lies the assumption that unseen target distributions could be covered by augmenting the mean and standard deviation of the low-level features. While the correlation between ``style'' and these parameters has been shown in previous research~\cite{pan2018two,zhou2021domain}, we believe that the hypothesis stating that the domain shift could be described only by these parameters over-simplifies generalization. Moreover, \method does not handle or provide an estimation of uncertainty, which is crucial for both classes in and outside the label set for the application at hand.

\subsection{Perspectives}
\label{sec:persp}
Vision transformers (ViTs)~\cite{dosovitskiy2021an} have recently emerged as an alternative to CNNs. We leave a ViT implementation of \method for future work. Applying prompt-driven instance normalization (PIN)~\cite{fahes2023poda} to ViTs appears non-trivial as the relation between statistics of low-level feature maps and style is established only for CNNs so far, and could raise some technical challenges. Exploring this direction might first involve a study of the correlation between style and statistics of patches. If such correlation is demonstrated, a naive way to apply \method could be by applying PIN with tied parameters across the patches.

While some modern architectures are inherently more robust than older ones, the problem of DGSS with ResNets (\eg ResNet-50 and ResNet-101) is still not solved. As long as the gap exists between in-domain and out-of-distribution performances, we believe that this setting remains interesting, and that a general understanding of domain generalization could emerge from the algorithms proposed to address it.

{
    \small
    \bibliographystyle{ieeenat_fullname}
    \bibliography{main}
}

\end{document}